\pdfoutput=1
\PassOptionsToPackage{table,dvipsnames,xcdraw}{xcolor}
\documentclass{article}
\usepackage{preprint,times}
\usepackage[T1]{fontenc}
\usepackage{microtype}

\usepackage{amsmath,amssymb,amsthm}
\usepackage{graphicx}
\usepackage{booktabs}
\usepackage{colortbl}
\usepackage{pifont}
\usepackage{enumitem}
\usepackage{needspace}
\usepackage{caption}
\usepackage{url}
\usepackage{xurl}
\usepackage{hyperref}
\usepackage[all]{hypcap}

\usepackage{nicefrac}
\usepackage{multirow}
\usepackage{cleveref}
\usepackage{dsfont}
\usepackage{makecell}
\usepackage{subcaption}
\usepackage{bbm}
\usepackage{wrapfig}
\usepackage{mathtools,xparse}
\usepackage{array}
\usepackage{arydshln}
\usepackage{scalerel}
\usepackage{tablefootnote}
\usepackage{adjustbox}
\usepackage{setspace}
\usepackage[most,skins,theorems]{tcolorbox}
\usepackage{titletoc}
\titlecontents{section}[1.8em]{\addvspace{7pt}\bfseries\color{cite_blue}}{\contentslabel{1.8em}}{}{\titlerule*[0.6pc]{.}\contentspage}
\titlecontents{subsection}[4.4em]{\addvspace{1pt}\color{cite_blue}}{\contentslabel{2.6em}}{}{\titlerule*[0.6pc]{.}\contentspage}

\usepackage{xspace}
\usepackage{pgfplots}
\usetikzlibrary{calc}
\pgfplotsset{compat=1.18}
\usepackage[normalem]{ulem}

\newtcolorbox{promptbox}{
  colback=gray!5,
  colframe=gray!70,
  boxrule=0.8pt,
  arc=2mm,
  left=10pt,
  right=10pt,
  top=10pt,
  bottom=10pt,
  width=\linewidth,
  fontupper=\ttfamily\footnotesize,
  enhanced,
}

\newboolean{showsection}
\setboolean{showsection}{true}
\makeatletter
\@namedef{ver@everyshi.sty}{}
\makeatother

\definecolor{custom_green}{rgb}{0.0, 0.5, 0.0}
\definecolor{custom_red}{rgb}{1.0, 0.01, 0.24}
\definecolor{custom_blue}{HTML}{C9DAF7}
\definecolor{custom_purple}{HTML}{D9D1E9}
\definecolor{cite_purple}{HTML}{7406a7}

\hypersetup{
    colorlinks = true,
    citecolor = {cite_blue},
    linkcolor = {cite_blue},
    urlcolor = {cite_blue},
    linktoc = all,
}

\newcommand{\eg}{\emph{e.g.},\ }
\newcommand{\ie}{\emph{i.e.},\ }

\definecolor{blanchedalmond}{rgb}{1.0, 0.92, 0.8}
\definecolor{carmine}{rgb}{0.59, 0.0, 0.09}
\definecolor{lightblue}{rgb}{0.22,0.45,0.70}%

\renewcommand{\mathbf}{\boldsymbol}

\makeatletter
\def\Ddots{\mathinner{\mkern1mu\raise\p@
\vbox{\kern7\p@\hbox{.}}\mkern2mu
\raise4\p@\hbox{.}\mkern2mu\raise7\p@\hbox{.}\mkern1mu}}
\makeatother

\numberwithin{equation}{section}

\definecolor{amaranth}{rgb}{0.9, 0.17, 0.31}
\definecolor{antiquebrass}{rgb}{0.8, 0.58, 0.46}
\definecolor{antiquefuchsia}{rgb}{0.57, 0.36, 0.51}
\definecolor{chromeyellow}{rgb}{0.31, 0.47, 0.26}

\newcommand{\1}{\mathds 1}

\usepackage{amsmath,amsfonts,bm}

\def\eqref#1{equation~\ref{#1}}

\def\1{\bm{1}}

\def\rva{{\mathbf{a}}}

\def\rvc{{\mathbf{c}}}

\def\rvo{{\mathbf{o}}}

\def\rvq{{\mathbf{q}}}

\DeclareMathAlphabet{\mathsfit}{\encodingdefault}{\sfdefault}{m}{sl}
\SetMathAlphabet{\mathsfit}{bold}{\encodingdefault}{\sfdefault}{bx}{n}

\def\gD{{\mathcal{D}}}

\makeatletter
\def\mathcolor#1#{\@mathcolor{#1}}
\def\@mathcolor#1#2#3{%
  \protect\leavevmode
  \begingroup
    \color#1{#2}#3%
  \endgroup
}
\makeatother

\Crefformat{equation}{#2Eq.\;(#1)#3}
\Crefformat{figure}{#2Figure #1#3}
\Crefformat{assumption}{#2Assumption #1#3}
\Crefname{assumption}{Assumption}{Assumptions}

\usepackage{crossreftools}
\graphicspath{{assets/}}

\definecolor{oursbg}{RGB}{221,235,247}
\definecolor{cornellred}{rgb}{0.7, 0.11, 0.11}
\definecolor{cadmiumgreen}{rgb}{0.0, 0.42, 0.24}
\definecolor{aliceblue}{rgb}{0.91, 0.94, 0.97}
\definecolor{darkblue}{rgb}{0.83, 0.89, 0.97}
\definecolor{Red7}{rgb}{0.941, 0.243, 0.243}
\definecolor{Green7}{RGB}{55, 178, 77}
\definecolor{Blue9}{rgb}{0.098,0.3,0.9}
\definecolor{SJViolet}{RGB}{105,100,171}
\definecolor{SJRed}{RGB}{237,109,107}
\definecolor{gapOurs}{HTML}{2A6FC9}
\definecolor{gapReal}{HTML}{1B1B1B}
\definecolor{gapMasq}{HTML}{7B3FB0}
\definecolor{gapEgo}{HTML}{E07B12}
\definecolor{gapHuman}{HTML}{8A8781}
\newcommand{\legsq}[1]{{\color{#1}\rule[-0.2ex]{1.5ex}{1.5ex}}}

\newcommand{\cmark}{\ding{51}}%
\newcommand{\xmark}{\ding{55}}%

\newlength{\thumbw}
\newcommand{\thumbat}[3]{%
  \IfFileExists{assets/#2}{\includegraphics[width=#1]{#2}}{%
    \fbox{\parbox[c][\dimexpr0.5625\dimexpr#1\relax-2\fboxsep-2\fboxrule\relax][c]{\dimexpr#1-2\fboxsep-2\fboxrule\relax}{%
      \centering\scriptsize\textcolor{black!45}{\textsf{#3}}}}}}

\newlength{\gapw}\newlength{\agw}\newlength{\embw}
\newcommand{\tinyaxis}[1]{{\scriptsize\itshape\textcolor{black!55}{(#1)}}}
\newcommand{\ours}{RoboTryOn\xspace}
\newcommand{\oursb}{\textbf{\ours (Ours)}\xspace}

\newcommand{\figorbox}[3]{%
  \IfFileExists{assets/#1}{\includegraphics[width=#2]{#1}}{%
    \fbox{\parbox[c][#3][c]{\dimexpr#2-2\fboxsep-2\fboxrule\relax}{\centering
      \textcolor{red}{\textbf{missing:}} \texttt{assets/#1}}}}}

\title{Synthesizing Dexterous Robot Demonstrations from Human Videos via Virtual Try-On}
\shorttitle{Synthesizing Dexterous Robot Demonstrations from Human Videos via Virtual Try-On}

\author{
Seungyong~Lee\textsuperscript{1}\ \
Suhyeok~Jang\textsuperscript{1,2}\ \
Minyoung~Kim\textsuperscript{3}\ \
Jinwoo~Shin\textsuperscript{1,2}
}
\affiliation{\textsuperscript{1}KAIST\quad \textsuperscript{2}RLWRLD\quad
\textsuperscript{3}Korea University}
\correspondence{Correspondence to: \href{mailto:sylee0802@kaist.ac.kr,jinwoos@kaist.ac.kr}{\{sylee0802, jinwoos\}@kaist.ac.kr}.
\\{\upshape\textbf{Project page}: \url{https://ryan-seungyong-lee.github.io/RoboTryOn}}}
\logo{kaist_ai_logo.pdf}

\begin{document}
\maketitle

\begin{abstract}
Robot foundation models have shown remarkable performance in robot control, yet adapting them to a new task still relies heavily on teleoperated demonstrations.
Egocentric human videos offer a scalable substitute, but overlaying rendered robots on human videos or retargeting human poses into robot actions leaves significant visual or action gaps for high-DoF robot embodiments.
To address this limitation, we propose \ours, which treats human-to-robot translation as virtual try-on: try-on keeps the person and swaps the garment, \ours keeps the demonstrated motion and swaps the human for the robot.
Specifically, \ours converts a human video into a video of the target robot specified by a reference image, while preserving the original task, objects, and background. To maintain visual consistency, we translate selected keyframes autoregressively and use them to guide video synthesis with motion conditioned on the human hand skeleton.
Additionally, we use an inverse dynamics model (IDM) that incorporates depth features to infer action labels from synthesized videos for policy learning.
Our experiments across multiple humanoid embodiments show that policies trained with our synthetic data improve the average success rate on novel tasks, including challenging bimanual manipulation, from 2\% to 60\%, without any real robot demonstration of those tasks.
Moreover, training with our synthetic data matches the performance of policies trained solely on teleoperated demonstrations at approximately a third of the collection cost, demonstrating the cost efficiency of our approach.

\end{abstract}
\vspace{-0.1in}

\section{Introduction}
\label{sec:intro}

Robot foundation models trained on large-scale demonstrations have made
substantial progress toward general-purpose robot control
\citep{oneill2024oxe,octo2024,kim2024openvla,black2024pi0,liu2024rdt,nvidia2025gr00t,intelligence2025pi05},
while extending them to new tasks and environments still requires additional
robot demonstrations
\citep{oneill2024oxe,khazatsky2024droid,bu2025agibot}. Collecting such
demonstrations through teleoperation requires robot hardware, a
teleoperation system, and substantial operator effort
\citep{zhao2023aloha,wu2024gello,chi2024umi,khazatsky2024droid}. The
burden is heaviest for dexterous hands, where operators must
coordinate many degrees of freedom and fine-grained contacts through
retargeting interfaces that are challenging to develop and operate
\citep{handa2020dexpilot,qin2023anyteleop,cheng2024television,wang2024dexcap,xu2025dexumi}.

Alternatively, human videos offer another way to demonstrate a new task.
In particular, egocentric recordings of human object manipulation already exist at internet scale and far outnumber the demonstrations teleoperation can produce, ranging from in-the-wild collections to manually annotated corpora recorded for robot learning~\citep{goyal2017something,shan2020100doh,damen2022epickitchens,grauman2022ego4d,grauman2024egoexo4d,liu2022hoi4d,hoque2025egodex,li2025vitra,punamiya2026egoverse}.
They capture how people grasp and manipulate objects, including coordinated bimanual interaction, without a robot during collection. Recent works use these videos to learn transferable visual and behavioral knowledge~\citep{nair2022r3m,yang2025egovla,luo2025beingh0,dyna2026dyna2}, or convert them into robot data through retargeting \citep{kareer2025egomimic}, rendering \citep{lepert2025phantom,lepert2025masquerade,wang2026ego2robot,liu2026egoengine}, or generative translation \citep{song2025mitty,ci2025h2rgrounder}.

However, a human video is separated from a robot demonstration by a
visual gap and an action gap, and both are severe for dexterous hands
\citep{liu2025immimic,punamiya2025egobridge,liu2026egoengine}.
In terms of appearance, human videos show a human arm and hand, whereas the policy observes a robot at deployment. Prior work narrows this gap by masking the human arm or
overlaying a rendered robot, and must apply the same edit to the
robot's camera stream during both training and deployment
\citep{kareer2025egomimic,lepert2025phantom,lepert2025masquerade}.
In terms of action, the reconstructed hand pose is itself imprecise, and no exact retargeting map takes it to robot joint commands, since human and robot hands differ in morphology and kinematics \citep{handa2020dexpilot,qin2023anyteleop,kong2026retargeting}.
Even a perfect mapping would reproduce only the shape of the hand, not how firmly the robot must grip the object \citep{chen2025dexforce,liu2026egoengine}.

\textbf{Our approach.}
We address these gaps by treating human-to-robot transfer as virtual
try-on, which keeps the motion and changes the embodiment. \ours
converts a human manipulation video into a paired robot demonstration
with action labels. To this end, we first generate a target robot video
performing the same task in the source scene and then predict the
corresponding joint-space actions with an inverse dynamics model (IDM),
rather than deriving robot actions directly from reconstructed hand
motion. As in keyframe-based video virtual try-on~\citep{zuo2025dreamvvt}, we
autoregressively translate selected human frames into robot keyframes,
and synthesize the motion between consecutive keyframes under the
guidance of the human hand skeleton flow.
Our generative models learn this translation from robot videos through a shared arm mask and hand skeleton interface, allowing the same conditions to be extracted from human videos at inference without paired human and robot demonstrations. We then train a policy on the synthesized robot demonstrations, enabling the robot to learn new tasks with few or no teleoperated demonstrations.

Our experiments show that \ours consistently improves policy
performance when task-specific robot demonstrations are limited or
unavailable. In particular, averaged over two robot embodiments, \ours
increases the success rate from 2.1\% to 60.0\% on out-of-domain tasks with goals, skills, and objects unseen in real robot
demonstrations. It also achieves 45.9\% on challenging bimanual tasks
despite all teleoperated demonstrations containing only single-arm
manipulation. Finally, training with
demonstrations derived from human video matches the performance of
training on teleoperated demonstrations alone while requiring approximately a third
of the collection cost (see \Cref{fig:cost}).

We summarize our main contributions below:
\begin{itemize}[leftmargin=*,topsep=0.0em, itemsep=0.0em]
\item We introduce \ours, which turns an egocentric human video into a
robot demonstration the way virtual try-on swaps a garment, with
generative models trained on robot data alone.
\item We show that our synthetic data bridge the visual and
action gaps in human videos, removing the need for test-time visual editing and generating better action labels than retargeted actions.
\item Through experiments on two real robots with dexterous hands over diverse out-of-domain tasks, we demonstrate that our synthesized demonstrations enable learning tasks without teleoperated demonstrations and reduce data collection costs relative to teleoperation.
\end{itemize}

\section{Related Work}
\label{sec:related}

\textbf{Learning from human videos.}
Human videos provide a scalable source of visual and behavioral
knowledge for robot learning.
Early work learns visual representations from human videos for downstream robot control \citep{nair2022r3m,majumdar2023vc1}, while recent approaches incorporate human videos into robot foundation model training to learn transferable visual and behavioral knowledge \citep{yang2025egovla,luo2025beingh0,dyna2026dyna2,nvidia2025gr00t}.
Another line converts human demonstrations into robot training data through motion retargeting \citep{kareer2025egomimic}, often combined with rendered robot observations
\citep{lepert2025phantom,lepert2025masquerade,wang2026ego2robot}.
Approaches based on simulation further construct digital twins to generate robot observations and refine action trajectories \citep{liu2026egoengine}.

\textbf{Generative human-to-robot translation.}
Recent advances in video diffusion models have enabled direct
translation of human demonstrations into robot videos. Mitty
\citep{song2025mitty} conditions generation on human videos and learns from
paired human and robot data, including synthetically rendered pairs.
H2R-Grounder \citep{ci2025h2rgrounder} removes the need for paired supervision
by reconstructing robot videos from inpainted backgrounds and
gripper pose cues. Our approach supports multiple dexterous embodiments through shared generative models and produces demonstrations with action labels for real-world policy learning.

\textbf{Video synthesis and action labeling.}
Constructing robot demonstrations through video generation requires
both temporally consistent synthesis and corresponding action labels.
Video virtual try-on studies consistency when modifying a subject's
appearance \citep{fang2024vivid,zuo2025dreamvvt}. In particular,
DreamVVT generates keyframes and uses them to guide video synthesis.
We adopt this decomposition for embodiment translation and, following prior work~\citep{baker2022vpt, du2023unipi,jang2025dreamgen}, use IDMs to infer actions from the resulting videos.
\vspace{0.1in}

\section{Method}
\label{sec:method}

\begin{figure}[t]
\centering
\vspace{-0.1in}
\figorbox{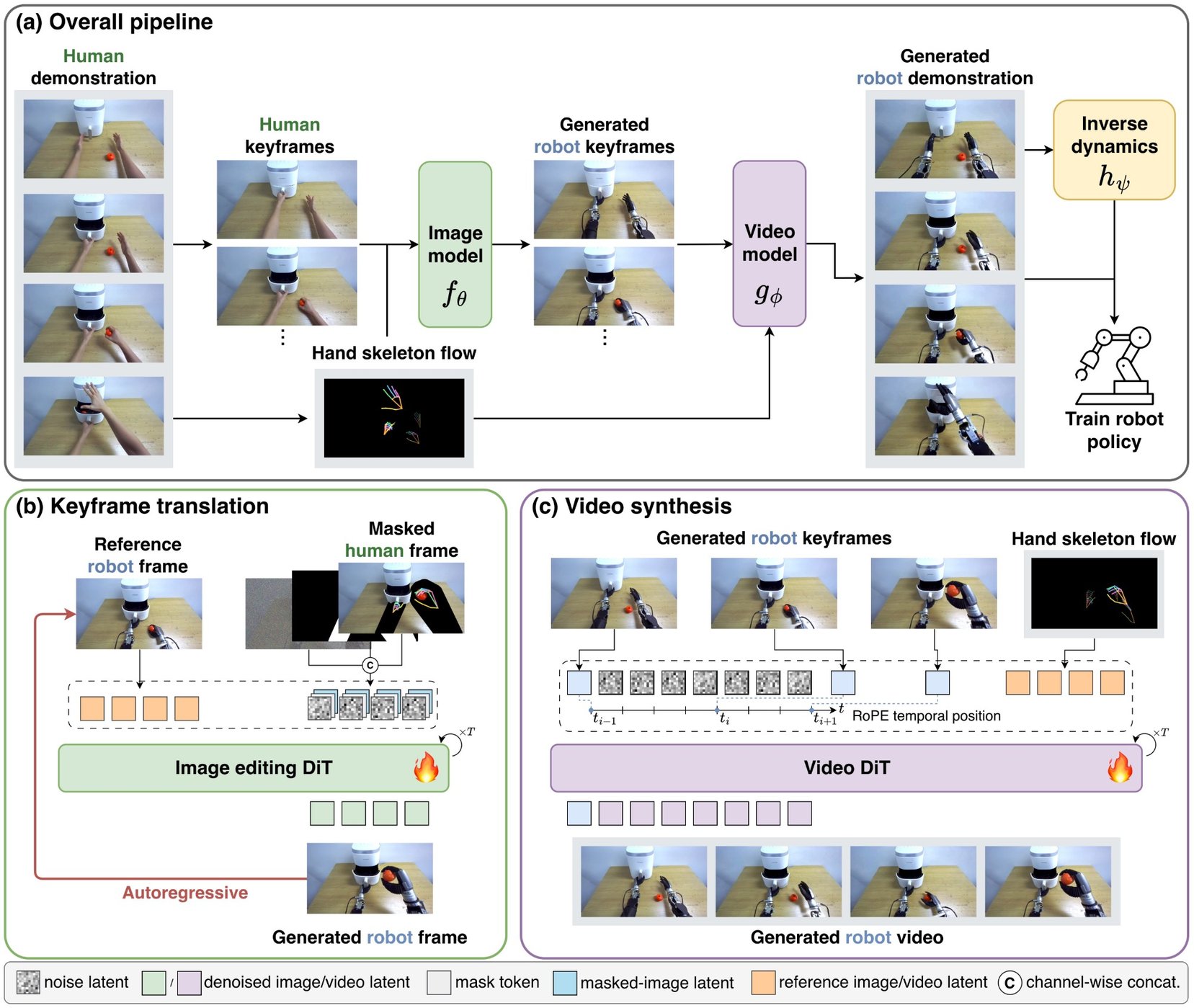}{0.9\textwidth}{4.5cm}
\caption{\textbf{\ours pipeline.} (a)~The image model sequentially translates keyframes
of a human video into robot keyframes, the video
model synthesizes the motion between them under the guidance of the
hand skeleton flow, and the inverse dynamics model labels the actions. (b)~The image model inpaints the masked human frame, using the
previous robot keyframe as a reference. (c)~The video model synthesizes
the robot video, conditioned on the robot keyframes at their temporal
positions and the skeleton flow.}
\label{fig:pipeline}
\vspace{-0.1in}
\end{figure}

\subsection{Problem Setup and Overview}
\label{sec:overview}

\textbf{Policy learning from demonstrations.} We consider a teleoperated robot demonstration dataset $\gD=\{\mathbf{X}^{(i)}\}_{i=1}^{N}$, where each demonstration $\mathbf{X}^{(i)}=\big((\rvo_t,\rvq_t,\rva_t)_{t=1}^{T^{(i)}}, \rvc\big)$ contains a visual observation $\rvo_t$, a proprioceptive state $\rvq_t$, and a robot action $\rva_t$ at each timestep $t\in\{1,\dots,T^{(i)}\}$, together with a task instruction $\rvc$. We mainly use vision-language-action models (VLAs) as our base policy $\pi_\omega$, which predicts an action chunk $\mathbf{A}_t=\rva_{t:t+H-1}$ of horizon $H$ conditioned on $(\rvo_t,\rvq_t,\rvc)$.

\textbf{Demonstrations from human video.}
\ours aims to augment $\gD$ with demonstrations synthesized from human video to address the expensive cost of collecting real robot data.
A human demonstration $\mathbf{X}^h=\big((\rvo^h_t)_{t=1}^{T},\rvc\big)$\footnote{For simplicity, we omit the demonstration index.} contains egocentric observations $\rvo^h_t$ and the instruction $\rvc$.
Given $\mathbf{X}^h$ and a reference image of the
target robot $r$, \ours uses a three-stage pipeline to generate a robot demonstration
$\hat{\mathbf{X}}^r=\big((\hat\rvo^r_t,\hat\rvq_t,\hat\rva_t)_{t=1}^{T},\rvc\big)$
that follows the human motion (see \Cref{fig:pipeline}).
First, an image generation model $f_\theta$ translates selected human keyframes into robot keyframes (\Cref{sec:keyframe}).
Next, a video generation model $g_\phi$ synthesizes the robot video between keyframes under the guidance of the hand skeleton (\Cref{sec:video}).
Finally, an inverse dynamics model $h_\psi$ predicts the action labels $\hat\rva$ from the synthesized video, and the states $\hat\rvq$ are then derived from the predicted actions (\Cref{sec:idm}).
The resulting demonstrations form synthetic data $\hat{\gD}$, on which we train the policy $\pi_\omega$ alone or together with $\gD$ (\Cref{sec:policy}).
We train the three models and the policy in the same way, as conditional flow-matching models~\citep{lipman2023flow,liu2023rectified}. We provide the details of the flow matching process in \Cref{app:flow}.

\subsection{Keyframe Translation}
\label{sec:keyframe}
In this section, we describe how \ours\ selects keyframes from the human demonstration $\mathbf{X}^h$ and translates them into target robot keyframes.
We first train the image generation model $f_\theta$ on robot demonstrations and apply it to the selected human keyframes at inference.

\textbf{Keyframes.}
We select keyframes from the wrist trajectories of the reconstructed human hands~\citep{zhang2025hawor}. A frame is selected when the speed of both wrists reaches a local minimum, where the hands are unblurred, and a wrist has moved more than a threshold from its position in the previous keyframe. We also always include the first and last frames. The resulting keyframe indices form the set $\mathcal{T}=\{t_1<\dots<t_L\}$.

\textbf{Conditioning.}
We condition $f_\theta$ on an arm and hand mask $\mathbf{M}_t$ and a hand skeleton $\mathbf{S}_t$. As in virtual try-on methods~\citep{choi2021vitonhd,choi2024idmvton,jiang2024fitdit,lee2025voost}, we enlarge and smooth $\mathbf{M}_t$, so its outline does not reveal the embodiment it covers. The skeleton $\mathbf{S}_t$ is a 2D projection of hand keypoints obtained by forward kinematics for a robot or by retargeting the reconstructed hand for a human. We define a conditioning function $\Phi(\rvo;\mathbf{M},\mathbf{S})\coloneqq\big((\mathbf{1}-\mathbf{M})\odot\rvo\big)\oplus\mathbf{S}$, which masks out the region $\mathbf{M}$ of a frame $\rvo$ and draws the skeleton $\mathbf{S}$ over it. The same function gives the conditioned robot frame $\tilde\rvo_t=\Phi(\rvo_t;\mathbf{M}_t,\mathbf{S}_t)$ for training and the conditioned human frame $\tilde\rvo^h_t=\Phi(\rvo^h_t;\mathbf{M}^h_t,\mathbf{S}^h_t)$ for inference. To make the model robust to imperfect hand retargeting, we perturb and randomly drop the skeleton during training. The skeleton then provides only coarse guidance on hand position and motion, while detailed hand configuration comes from the learned prior over robot hands.

\textbf{Model.}
The image generation model $f_\theta$ aims to translate a human keyframe into a robot
keyframe as inpainting guided by a reference image: it fills the masked region $\mathbf{M}_t$ with the target robot in the pose specified by the skeleton $\mathbf{S}_t$, using the reference image to determine its appearance (see~\Cref{fig:pipeline}(b)).

We train $f_\theta$ to generate a robot frame $\rvo_t$ from two conditions: the conditioned frame $\tilde\rvo_t$ with its mask $\mathbf{M}_t$, and a reference frame $\rvo_{t'}$ from the same episode at an earlier time $t'<t$, which provides the appearance of the embodiment, the objects, and the background. With a pretrained image encoder $\mathcal{E}_{\mathrm{img}}$ and $\mathbf{z}=\mathcal{E}_{\mathrm{img}}(\rvo_t)$, we train $f_\theta$ using the flow-matching objective with diffusion timestep $\tau$:
\begin{equation}
    \mathcal{L}_{\mathrm{img}}(\theta)
    =
    \mathbb{E}_{(\rvo_t,\, \rvo_{t'}) \sim \gD,\; \boldsymbol{\epsilon},\; \tau}
    \left[
        \left\|
            f_\theta\big(\mathbf{z}^{\tau}, \tau \mid
            \tilde\rvo_t,\; \mathbf{M}_t,\; \rvo_{t'}\big)
            - (\mathbf{z} - \boldsymbol{\epsilon})
        \right\|_2^2
    \right].
    \label{eq:img}
\end{equation}
At inference, $f_\theta$ translates the human keyframes in temporal order, each conditioned on the previously translated robot keyframe:
\begin{equation}
    \hat\rvo^r_{t_j} \sim p_\theta\big(\, \cdot \mid
    \tilde\rvo^h_{t_j},\; \mathbf{M}^h_{t_j},\; \hat\rvo^r_{t_{j-1}} \big),
    \qquad j = 1, \dots, L,
    \label{eq:ar}
\end{equation}
where $\hat\rvo^r_{t_0}$ is a reference image of the target robot, which selects the embodiment.

\subsection{Video Synthesis}
\label{sec:video}

The video generation model $g_\phi$ interpolates between the
translated keyframes $\hat\rvo^r_{\mathcal{T}}=(\hat\rvo^r_t)_{t\in\mathcal{T}}$ to synthesize the full robot video, guided by the hand skeleton flow $\mathbf{S}_{1:T}=(\mathbf{S}_t)_{t=1}^{T}$ (see \Cref{fig:pipeline}(c)).
During training, $g_\phi$ generates a robot video $\rvo_{1:T}$ conditioned on its keyframes $\rvo_{\mathcal{T}}$ and its skeleton flow $\mathbf{S}_{1:T}$, whose keypoints are perturbed and dropped as for $f_\theta$. With a pretrained video encoder $\mathcal{E}_{\mathrm{vid}}$ and $\mathbf{w} = \mathcal{E}_{\mathrm{vid}}(\rvo_{1:T})$, the objective is
\begin{equation}
    \mathcal{L}_{\mathrm{vid}}(\phi)
    =
    \mathbb{E}_{\rvo_{1:T} \sim \gD,\; \boldsymbol{\epsilon},\; \tau}
    \left[
        \left\|
            g_\phi\big(\mathbf{w}^{\tau}, \tau \mid
            \rvo_{\mathcal{T}},\; \mathbf{S}_{1:T}\big)
            - (\mathbf{w} - \boldsymbol{\epsilon})
        \right\|_2^2
    \right].
    \label{eq:vid}
\end{equation}%
At inference, $g_\phi$ synthesizes the robot video $\hat\rvo^r_{1:T}$ conditioned on $(\hat\rvo^r_{\mathcal{T}},\mathbf{S}^h_{1:T})$, as follows:
\begin{equation}
    \hat\rvo^r_{1:T} \sim p_\phi\big(\, \cdot \mid
    \hat\rvo^r_{\mathcal{T}},\; \mathbf{S}^h_{1:T} \big).
    \label{eq:video}
\end{equation}%
If $T$ exceeds the maximum length that $g_\phi$ can generate at once, we synthesize the video as consecutive clips split at keyframes, where adjacent clips share the keyframe at their boundary.

\subsection{Pseudo Action Labeling}
\label{sec:idm}

The inverse dynamics model $h_\psi$ infers robot action labels $\hat\rva_{1:T}$ from the synthesized video $\hat\rvo^r_{1:T}$. From two observations $K$ steps apart, $h_\psi$ predicts the chunk of joint-space actions between them. We train $h_\psi$ using actions from teleoperated robot demonstrations as targets:
\begin{equation}
    \mathcal{L}_{\mathrm{idm}}(\psi)
    =
    \mathbb{E}_{(\rvo_t,\, \rvo_{t+K},\, \rva_{t:t+K}) \sim \gD,\; \boldsymbol{\epsilon},\; \tau}
    \left[
        \left\|
            h_\psi\big(\rva^{\tau}_{t:t+K}, \tau \mid
            \rvo_t,\; \rvo_{t+K}\big)
            - (\rva_{t:t+K} - \boldsymbol{\epsilon})
        \right\|_2^2
    \right].
    \label{eq:idm}
\end{equation}
At inference, $h_\psi$ labels the synthesized video as
$\hat\rva_{t:t+K} \sim p_\psi(\,\cdot \mid \hat\rvo^r_t, \hat\rvo^r_{t+K})$
over the whole sequence. The states are then derived from the action labels as $\hat\rvq_t=\hat\rva_{t-d}$, where $d$ is the lag between the commanded action and the recorded state, measured on the teleoperated data.

Since the IDM estimates wrist positions less accurately than the hand reconstruction, we use inverse kinematics to correct the wrist position in the action labels while keeping the hand joints predicted by the IDM. We provide further implementation details for all three models in \Cref{app:robotryon}.

\subsection{Policy Training}
\label{sec:policy}

Each synthesized demonstration $\hat{\mathbf{X}}^r$ has the same form as a teleoperated one, and the demonstrations synthesized from all human videos form $\hat\gD$. We train the policy on $\gD\cup\hat\gD$, or on $\hat\gD$ alone.

\textbf{Training objective and inference.}
Following recent VLAs~\citep{black2024pi0,intelligence2025pi05,nvidia2025gr00t}, we fine-tune the pretrained VLA as follows:
\begin{equation}
    \mathcal{L}_{\mathrm{policy}}(\omega)
    =
    \mathbb{E}_{(\rvo_t, \rvq_t, \mathbf{A}_t, \rvc) \sim \gD\cup\hat\gD,\; \boldsymbol{\epsilon},\; \tau}
    \left[
        \left\|
            \pi_\omega\big(\mathbf{A}_t^{\tau}, \tau \mid
            \rvo_t, \rvq_t, \rvc\big)
            - (\mathbf{A}_t - \boldsymbol{\epsilon})
        \right\|_2^2
    \right].
    \label{eq:policy}
\end{equation}
At inference, $\pi_\omega$ generates the action chunk $\mathbf{A}_t$ from Gaussian noise conditioned on $(\rvo_t,\rvq_t,\rvc)$ with Euler ODE sampling~\citep{lipman2023flow}.

\section{Experiments}
\label{sec:experiments}

We validate \ours on real robots with dexterous hands and investigate
the following questions:
\begin{itemize}[leftmargin=*,itemsep=0mm]
\item Can a policy learn tasks never demonstrated through teleoperation from
human video alone, and does \ours outperform other retargeting or overlaying methods? (\Cref{tab:main_real})
\item Does our synthesized data bridge the visual and action gaps between human videos and real robot data? (\Cref{fig:visual_gap,fig:action_gap,fig:action_vocab})
\item Can synthesized data replace teleoperated
demonstrations, and how does the success rate change with the amount of human
video? (\Cref{fig:cost})
\end{itemize}

\subsection{Experimental Setup}
\label{sec:setup}

\begin{figure}[t]
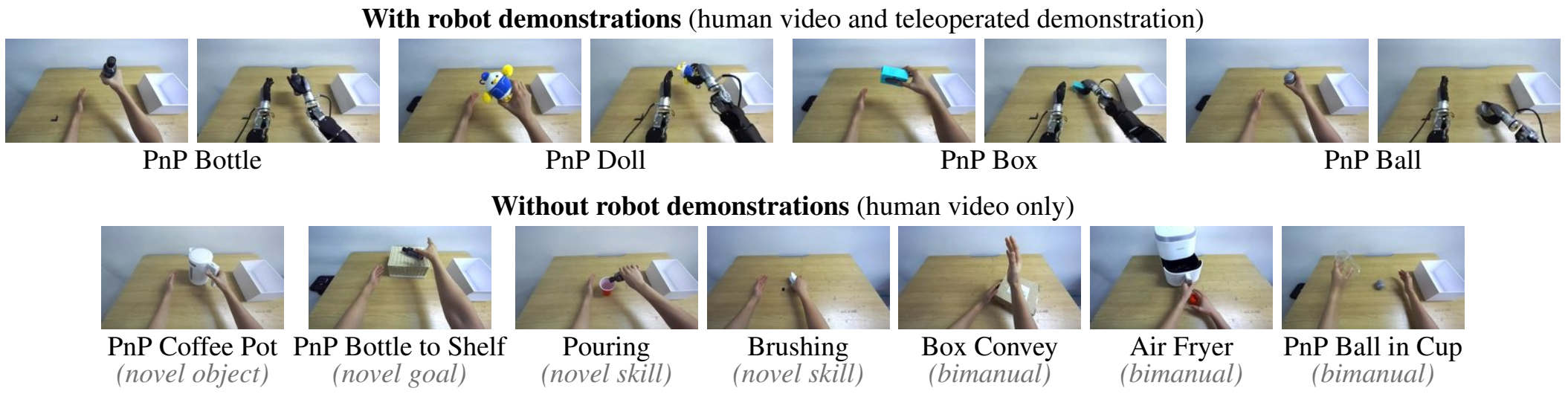

\centering
\setlength{\tabcolsep}{1.2pt}
\renewcommand{\arraystretch}{0.8}
\newcommand{\gt}[2]{\thumbat{0.116\textwidth}{#1}{#2}}
\scriptsize
\begin{tabular}{@{}cc@{\hspace{5pt}}cc@{\hspace{5pt}}cc@{\hspace{5pt}}cc@{}}
\multicolumn{8}{c}{\textbf{With robot demonstrations} (human video and teleoperated demonstration)} \\[1pt]
\gt{tasks/human-adj/pnp-bottle.jpg}{bottle / human} & \gt{tasks/teleop-adj/pnp-bottle.jpg}{bottle / robot} &
\gt{tasks/human-adj/pnp-doll.jpg}{doll / human} & \gt{tasks/teleop-adj/pnp-doll.jpg}{doll / robot} &
\gt{tasks/human-adj/pnp-box.jpg}{box / human} & \gt{tasks/teleop-adj/pnp-box.jpg}{box / robot} &
\gt{tasks/human-adj/pnp-ball.jpg}{ball / human} & \gt{tasks/teleop-adj/pnp-ball.jpg}{ball / robot} \\
\multicolumn{2}{c}{PnP Bottle} & \multicolumn{2}{c}{PnP Doll} & \multicolumn{2}{c}{PnP Box} & \multicolumn{2}{c}{PnP Ball} \\
\end{tabular}

\vspace{4pt}
\begin{tabular}{@{}ccccccc@{}}
\multicolumn{7}{c}{\textbf{Without robot demonstrations} (human video only)} \\[1pt]
\gt{tasks/human-adj/pnp-coffee-pot.jpg}{coffee pot} & \gt{tasks/human-adj/bottle-on-shelf.jpg}{bottle to shelf} &
\gt{tasks/human-adj/pouring.jpg}{pouring} & \gt{tasks/human-adj/brush.jpg}{brushing} &
\gt{tasks/human-adj/box_convey.jpg}{box convey} & \gt{tasks/human-adj/air-fryer.jpg}{air fryer} & \gt{tasks/human-adj/ball-in-cup.jpg}{ball in cup} \\
PnP Coffee Pot & PnP Bottle to Shelf & Pouring & Brushing & Box Convey & Air Fryer & PnP Ball in Cup \\
\tinyaxis{novel object} & \tinyaxis{novel goal} & \tinyaxis{novel skill} & \tinyaxis{novel skill} & \tinyaxis{bimanual} & \tinyaxis{bimanual} & \tinyaxis{bimanual} \\
\end{tabular}
\caption{\textbf{Examples of real-world benchmark tasks.} For each of the four PnP tasks (top), both a human video
and a teleoperated demonstration are available. For each of the remaining seven tasks (bottom), only a human video is available.}
\label{fig:tasks}
\vspace{-0.2in}
\end{figure}

\textbf{Robots and tasks.}
We evaluate \ours on two bimanual robots with dexterous hands: an
OpenArm with two 7-DoF arms and two 6-DoF Inspire RH56F1 hands, and an
RB-Y1 with two 7-DoF arms and two 20-DoF WujiHand2 hands. Both are
evaluated on eleven tabletop tasks (see \Cref{fig:tasks}). There are four pick-and-place (PnP) tasks, \ie in-domain tasks, with 50 teleoperated demonstrations per robot
and 50 human videos each: \emph{PnP Bottle}, \emph{Doll}, \emph{Box},
and \emph{Ball}. The other seven tasks, \ie out-of-domain tasks, have 50 human videos each and no
robot demonstrations, and they introduce a single novel
object task (\emph{PnP Coffee Pot}), a single novel goal
task (\emph{PnP Bottle to Shelf}), two novel skill tasks (\emph{Pouring},
\emph{Brushing}), and three bimanual manipulation tasks (\emph{Box Convey},
\emph{Air Fryer}, \emph{PnP Ball in Cup}), whereas the teleoperated tasks are all
single-arm tasks. Further details on the robots and tasks are
provided in \Cref{app:env}.

\textbf{Baselines.}
We compare with the two existing ways of turning human video into robot
training data. Both retarget the reconstructed hand into actions and
differ in how they handle the human in the image.
\emph{EgoMimic} \citep{kareer2025egomimic} masks the arm out of the
human frames, while \emph{Masquerade} \citep{lepert2025masquerade}
inpaints the arm out, composites a rendering of the robot in the
retargeted pose, and adds an auxiliary loss on future 2D robot keypoints.
Following their protocols, both apply the same edit to the robot data
and to the live camera image at test time, whereas \ours runs on the raw
camera view (see details in \Cref{app:baselines}). All methods share the
policy model, the teleoperated episodes, and the human videos, so they
differ only in how the human video is converted.

\textbf{Implementation details.}
For each robot, we fine-tune $\pi_{0.5}$~\citep{intelligence2025pi05} for the same number of training steps on 200 teleoperated episodes (50 per in-domain task), on 550 episodes synthesized by \ours from human videos (50 per task over all eleven tasks), or on all 750 together. We evaluate each policy over 28 trials per task with randomized object positions (see \Cref{app:eval}). We train the image generation model, the video generation model, and the IDM on the teleoperated demonstrations of the four in-domain tasks and on additional robot videos, which enrich the embodiment prior of the models. The additional videos are not used to train any policy. We provide further details on the backbones and training in \Cref{app:trsetup}.

\begin{table*}[t]
\centering
\vspace{-0.1in}
\caption{
\textbf{Real-world results on (a) OpenArm and (b) RB-Y1.} Success
rate (\%) over 28 trials per task. In-domain: the four PnP tasks with
robot demonstrations. Out-of-domain: the seven tasks with human video
only, grouped by what they add. ``+'' rows add the converted human
videos to the teleoperated demonstrations, and the last row uses the
synthesized episodes alone. Baselines are evaluated following their own
protocols. Per-task results for the out-of-domain tasks are provided in
\Cref{app:extra}.
}
\label{tab:main_real}

\begin{subtable}{\textwidth}
\centering
\caption{OpenArm Benchmark Results.}
\label{tab:main_openarm}
\footnotesize
\begin{adjustbox}{max width=\textwidth}
\begin{tabular}{l cc c ccccc}
\toprule
\multirow{2.5}{*}{Method}
& \multicolumn{2}{c}{Data Source}
& \multirow{2.5}{*}{In-domain}
& \multicolumn{5}{c}{Out-of-domain} \\
\cmidrule(lr){2-3}
\cmidrule(lr){5-9}
& Robot & Human
& & Novel Object & Novel Goal & Novel Skill & Bimanual & Avg. \\
\midrule
Teleop only
& \textcolor{SJViolet}{\cmark} & \textcolor{SJRed}{\xmark}
& 47.3 & 14.3 & 7.1 & 0.0 & 0.0 & 3.1 \\
+ EgoMimic~\citep{kareer2025egomimic}
& \textcolor{SJViolet}{\cmark} & \textcolor{SJViolet}{\cmark}
& 50.0 & 14.3 & 0.0 & 10.7 & 0.0 & 5.1 \\
+ Masquerade~\citep{lepert2025masquerade}
& \textcolor{SJViolet}{\cmark} & \textcolor{SJViolet}{\cmark}
& 67.9 & 21.4 & 7.1 & 14.3 & 0.0 & 8.2 \\
+ \oursb{}
& \textcolor{SJViolet}{\cmark} & \textcolor{SJViolet}{\cmark}
& \textbf{75.9} & \textbf{64.3} & \textbf{89.3} & \textbf{83.9}
& \textbf{63.1} & \textbf{73.0} \\
\midrule
\oursb{}
& \textcolor{SJRed}{\xmark} & \textcolor{SJViolet}{\cmark}
& 57.1 & 50.0 & 57.1 & 58.9 & 38.1 & 48.5 \\
\bottomrule
\end{tabular}
\end{adjustbox}
\end{subtable}

\medskip

\begin{subtable}{\textwidth}
\centering
\caption{RB-Y1 Benchmark Results.}
\label{tab:main_rby1}
\footnotesize
\begin{adjustbox}{max width=\textwidth}
\begin{tabular}{l cc c ccccc}
\toprule
\multirow{2.5}{*}{Method}
& \multicolumn{2}{c}{Data Source}
& \multirow{2.5}{*}{In-domain}
& \multicolumn{5}{c}{Out-of-domain} \\
\cmidrule(lr){2-3}
\cmidrule(lr){5-9}
& Robot & Human
& & Novel Object & Novel Goal & Novel Skill & Bimanual & Avg. \\
\midrule
Teleop only
& \textcolor{SJViolet}{\cmark} & \textcolor{SJRed}{\xmark}
& 69.6 & 7.1 & 0.0 & 0.0 & 0.0 & 1.0 \\
+ EgoMimic~\citep{kareer2025egomimic}
& \textcolor{SJViolet}{\cmark} & \textcolor{SJViolet}{\cmark}
& 63.4 & 14.3 & 0.0 & 10.7 & 0.0 & 5.1 \\
+ Masquerade~\citep{lepert2025masquerade}
& \textcolor{SJViolet}{\cmark} & \textcolor{SJViolet}{\cmark}
& 63.4 & 21.4 & 7.1 & 12.5 & 0.0 & 7.7 \\
+ \oursb{}
& \textcolor{SJViolet}{\cmark} & \textcolor{SJViolet}{\cmark}
& \textbf{79.5} & 60.7 & \textbf{67.9} & \textbf{57.1} & \textbf{28.6} & \textbf{46.9} \\
\midrule
\oursb{}
& \textcolor{SJRed}{\xmark} & \textcolor{SJViolet}{\cmark}
& 60.7 & \textbf{64.3} & 60.7 & 53.6 & 16.7 & 40.3 \\
\bottomrule
\end{tabular}
\end{adjustbox}
\end{subtable}
\vspace{-0.2in}
\end{table*}

\subsection{Main Results}
\label{sec:main}

\textbf{In-domain tasks.}
As shown in \Cref{tab:main_real}, \ours improves the success rate over both baselines on the in-domain tasks of both robots. In particular, the success rate rises from 47.3\% to 75.9\% on the OpenArm and from 69.6\% to 79.5\% on the RB-Y1. This suggests that synthesized demonstrations add useful variation even to tasks with teleoperated demonstrations, as they cover more object placements and more varied arm and hand motions than the 50 teleoperated demonstrations per task.

\textbf{Out-of-domain tasks.}
\ours yields the largest gains on the seven tasks without robot demonstrations, enabling the policy to learn these tasks, whereas both baselines largely fail with success rates below 10\% (see \Cref{tab:main_real}). In particular, the success rate rises from 3.1\% to 73.0\% on the OpenArm and from 1.0\% to 46.9\% on the RB-Y1. Notably, the policy also learns the bimanual tasks, although every teleoperated demonstration uses a single arm. This demonstrates that \ours turns human videos into robot demonstrations from which a policy learns new objects, goals, skills, and bimanual coordination. Moreover, a single image model, video model, and IDM serve both robots despite their different kinematics, hands, and camera placements (see \Cref{fig:embodiments}). Additionally, we analyze why the baselines fail in \Cref{sec:gap}.

\subsection{Analysis}
\label{sec:gap}

In this section, we analyze the quality of our synthesized data and the effects of human and translated robot data on policy learning, focusing on the amount of data, collection cost, and execution speed. Unless otherwise stated, we use $\pi_{0.5}$ fine-tuned on the OpenArm in these experiments.

\textbf{Visual gap.}
\label{sec:visual}
\begin{figure}[t]
\centering
\vspace{-0.0in}
\newsavebox{\gapabox}
\setlength{\gapw}{0.138\textwidth}
\setlength{\tabcolsep}{0.8pt}
\renewcommand{\arraystretch}{0.8}
\newcommand{\gp}[2]{\thumbat{\gapw}{visual-gap-adj/#1}{#2}}
\newcommand{\gh}[1]{\makebox[\gapw][c]{\scriptsize #1}}
\sbox{\gapabox}{\footnotesize\begin{tabular}{@{}ccccc@{}}
\gh{Human} & \gh{EgoMimic} & \gh{Masquerade} & \gh{\textbf{\ours (Ours)}} & \gh{Real robot} \\[1pt]
\gp{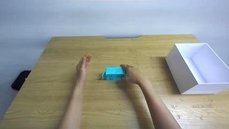}{h1} & \gp{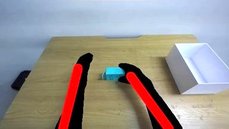}{e1} & \gp{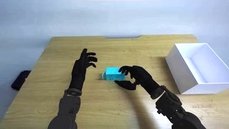}{m1} & \gp{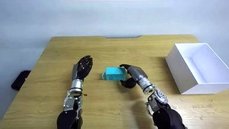}{r1} & \gp{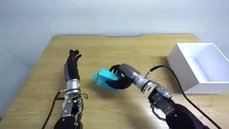}{r1} \\
\gp{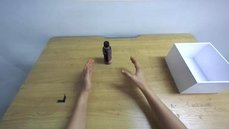}{h2} & \gp{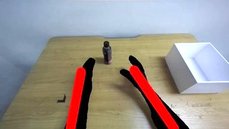}{e2} & \gp{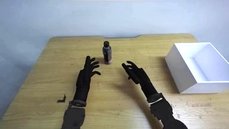}{m2} & \gp{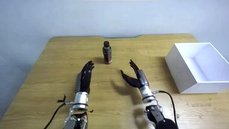}{r2} & \gp{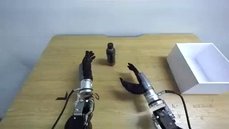}{r2} \\
\gp{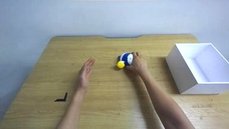}{h3} & \gp{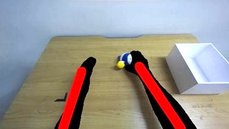}{e3} & \gp{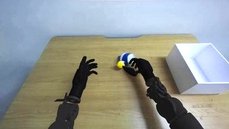}{m3} & \gp{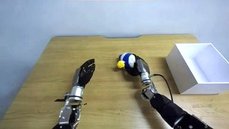}{r3} & \gp{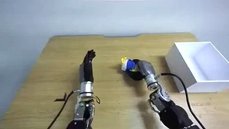}{r3} \\
\end{tabular}}
\begin{minipage}[b]{0.71\textwidth}
\centering
\usebox{\gapabox}\\[2pt]
{\footnotesize (a) Qualitative comparison}
\end{minipage}\hfill
\begin{minipage}[b]{0.28\textwidth}
\centering
\includegraphics[height=\dimexpr\ht\gapabox+\dp\gapabox-29pt\relax,width=\linewidth,keepaspectratio]{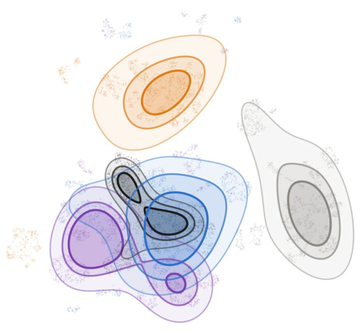}\\[2pt]
{\scriptsize
\legsq{gapHuman}\,Human\, \legsq{gapEgo}\,EgoMimic\, \legsq{gapMasq}\,Masquerade\\[1pt]
\legsq{gapOurs}\,\textbf{\ours (Ours)}\; \legsq{gapReal}\,Real robot}\\[2pt]
{\footnotesize (b) Feature distribution}
\end{minipage}
\caption{\textbf{Visual gap to real robot video.} (a)~The same moments as
recorded, as EgoMimic and Masquerade edit them, and as \ours translates
them, next to similar moments from teleoperation. (b)~UMAP of SigLIP features of
$\pi_{0.5}$ with density contours, using equal frame counts per corpus, real
robot frames from the four PnP tasks, and the other frames from all eleven tasks.}
\label{fig:visual_gap}
\end{figure}

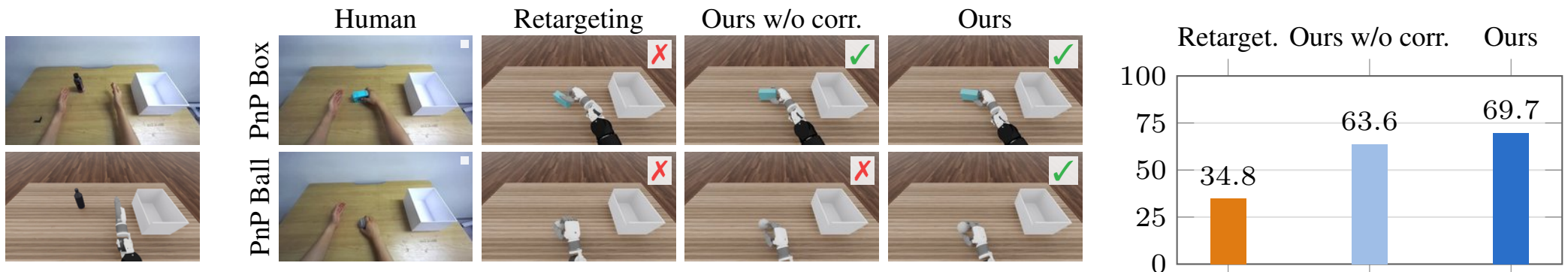
\begin{figure}[t]
\centering
\setlength{\agw}{0.122\textwidth}
\newcommand{\agh}{\dimexpr0.5625\agw\relax}%
\newcommand{\rowlab}[1]{\rotatebox{90}{\makebox[\agh][c]{#1}}}
\newcommand{\rtg}{Retargeting}
\newcommand{\okmark}{\textcolor{Green7}{\cmark}}
\newcommand{\nomark}{\textcolor{Red7}{\xmark}}
\newcommand{\agt}[3]{\begin{tikzpicture}[inner sep=0]
  \node[anchor=south west] (im) {\thumbat{\agw}{#1}{#2}};
  \node[anchor=north east, fill=white, fill opacity=0.85, text opacity=1,
        inner sep=1pt, font=\bfseries] at ([xshift=-1pt,yshift=-1pt]im.north east) {#3};
\end{tikzpicture}}
\scriptsize
\setlength{\tabcolsep}{1.2pt}
\renewcommand{\arraystretch}{0.8}
\begin{minipage}[b]{0.155\textwidth}
\centering
\begin{tabular}{@{}c@{}}
\vphantom{\rtg} \\
\thumbat{\agw}{action-gap-adj/real.jpg}{real} \\
\thumbat{\agw}{action-gap/sim.png}{sim} \\
\end{tabular}\\[2pt]
{\footnotesize (a) Real to sim}
\end{minipage}\hfill
\begin{minipage}[b]{0.54\textwidth}
\centering
\begin{tabular}{@{}l cccc@{}}
& \makebox[\agw][c]{Human} & \makebox[\agw][c]{\rtg} & \makebox[\agw][c]{Ours w/o corr.} & \makebox[\agw][c]{Ours} \\
\rowlab{PnP Box} &
\agt{action-gap-adj/ep1-human.jpg}{human}{} & \agt{action-gap/ep1-retarget.jpg}{retarget}{\nomark} & \agt{action-gap/ep1-idm.jpg}{idm}{\okmark} & \agt{action-gap/ep1-ours.jpg}{ours}{\okmark} \\
\rowlab{PnP Ball} &
\agt{action-gap-adj/ep2-human.jpg}{human}{} & \agt{action-gap/ep2-retarget.jpg}{retarget}{\nomark} & \agt{action-gap/ep2-idm.jpg}{idm}{\nomark} & \agt{action-gap/ep2-ours.jpg}{ours}{\okmark} \\
\end{tabular}\\[2pt]
{\footnotesize (b) Grasp moment}
\end{minipage}\hfill
\begin{minipage}[b]{0.29\textwidth}
\centering
\begin{tikzpicture}[baseline=(clabel.base)]
\begin{axis}[
    x=0.7307\agw, height=1.900\agw, ybar, bar width=9pt,
    ymin=0, ymax=100, ytick={0,25,50,75,100},
    tick label style={font=\scriptsize},
    symbolic x coords={R,I,C}, xtick={R,I,C},
    xticklabels={Retarget., Ours w/o corr., Ours},
    x tick label style={font=\scriptsize, text height=1.5ex, text depth=0.25ex}, xticklabel pos=upper,
    enlarge x limits={abs=0.45cm}, axis line style={black!60}, ymajorgrids, grid style={black!12},
    nodes near coords, nodes near coords style={font=\scriptsize, anchor=south, yshift=1pt},
    clip=false,
]
\addplot[fill=gapEgo, draw=none, bar shift=0pt] coordinates {(R, 34.8)};
\addplot[fill=gapOurs!45, draw=none, bar shift=0pt] coordinates {(I, 63.6)};
\addplot[fill=gapOurs, draw=none, bar shift=0pt] coordinates {(C, 69.7)};
\node[name=clabel, anchor=north, inner sep=0pt, font=\footnotesize] at ([yshift=-4pt]current axis.south) {(c) Replay success (\%)};
\end{axis}
\end{tikzpicture}
\end{minipage}
\caption{\textbf{Action gap: open-loop replay.} (a)~A real scene and its simulated replica.
(b)~The grasp moment of one human video under the three action
labelings, replayed open loop. (c)~Replay success over 66 episodes of three PnP
tasks (per task in \Cref{tab:action}).}
\label{fig:action_gap}
\vspace{-0.1in}
\end{figure}

We first examine whether synthesized videos resemble real robot videos (see \Cref{fig:visual_gap}). We encode the frames of raw, masked (EgoMimic), overlaid (Masquerade), and synthesized human videos with the visual encoder of the policy and compare their embeddings with those of real robot videos. Videos synthesized by \ours overlap with real robot videos in this embedding space, whereas raw, masked, and overlaid videos form separate clusters. Quantitatively, the Fr\'echet SigLIP distance~\citep{heusel2017gans,zhai2023siglip} to real robot videos agrees: \ours reduces it from 125.5 to 32.7, while overlay reduces it by about half and masking leaves it largely unchanged. This suggests that \ours closes the visual gap while the policy uses raw camera observations at test time. In contrast, both baselines narrow the gap only by applying the same masking or overlay to every camera observation during deployment.

\textbf{Action gap.}
\label{sec:action}
To evaluate action labels without ground truth robot actions, we replay them open loop in a simulated replica of each scene and measure task completion (see \Cref{fig:action_gap}a). We compare the labels from our IDM with retargeted labels, which map the reconstructed human hand to the robot hand~\citep{qin2023anyteleop} and solve for the arm by inverse kinematics. Replay succeeds in 69.7\% of episodes with our labels and 34.8\% with retargeted labels (see \Cref{fig:action_gap}c), and the difference arises mainly during grasping: retargeted motions bring the robot hand near the object, but imprecise hand reconstruction and the difference in hand geometry prevent a secure grasp (see \Cref{fig:action_gap}b). This suggests that labeling synthesized robot videos with the IDM yields more executable actions than retargeting, since \ours uses the reconstructed hand only to guide generation and the generated robot hand grasps the object as in the robot data.

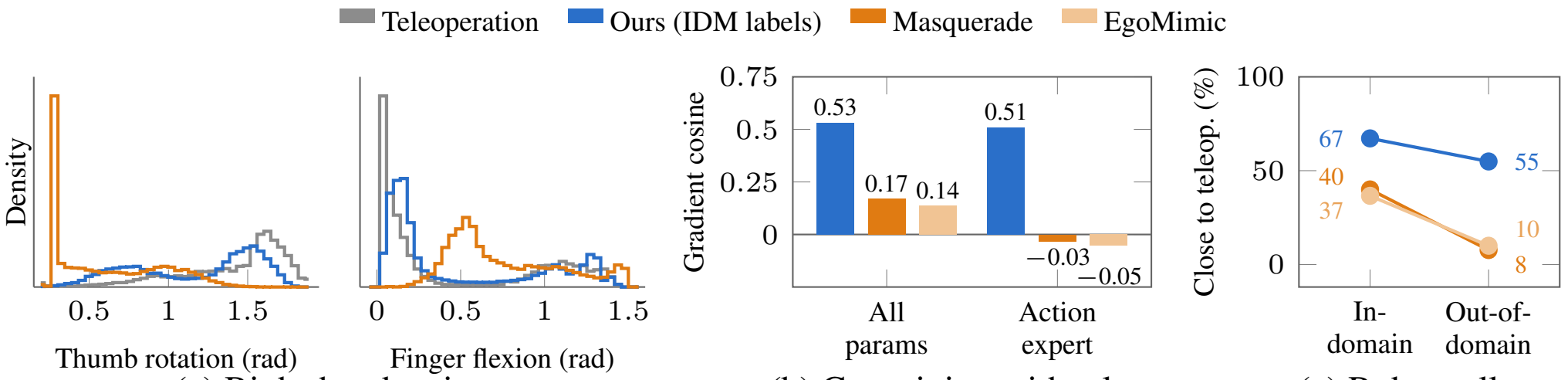
\begin{figure}[t]
\centering
{\scriptsize
\tikz[baseline=-0.6ex]\fill[black!45] (0,0) rectangle (0.3,0.12);~Teleoperation\quad
\tikz[baseline=-0.6ex]\fill[gapOurs] (0,0) rectangle (0.3,0.12);~Ours (IDM labels)\quad
\tikz[baseline=-0.6ex]\fill[gapEgo] (0,0) rectangle (0.3,0.12);~Masquerade\quad
\tikz[baseline=-0.6ex]\fill[gapEgo!45] (0,0) rectangle (0.3,0.12);~EgoMimic}\\[3pt]
\begin{tikzpicture}
\pgfplotsset{
  avbase/.style={scale only axis, height=1.8cm, axis line style={black!60},
    tick label style={font=\scriptsize}, label style={font=\scriptsize}, clip=false},
  avhist/.style={avbase, width=0.175\textwidth, ymin=0,
    ytick=\empty, axis y line*=left, axis x line*=bottom},
  avline/.style={const plot mark mid, no marks, line width=0.8pt},
}
\begin{axis}[avhist, name=thumb, xlabel={Thumb rotation (rad)}, ylabel={Density}, xmin=0.15, xmax=1.95, xtick={0.5,1,1.5}, ylabel style={yshift=-3pt}]
\addplot[avline, black!45] table[col sep=comma, x=bin_center_rad, y=teleop]{figures/data/action_gap/a_hist_rhand_thumb1_all12.csv};
\addplot[avline, gapOurs] table[col sep=comma, x=bin_center_rad, y=idm]{figures/data/action_gap/a_hist_rhand_thumb1_all12.csv};
\addplot[avline, gapEgo] table[col sep=comma, x=bin_center_rad, y=retarget]{figures/data/action_gap/a_hist_rhand_thumb1_all12.csv};
\end{axis}
\begin{axis}[avhist, name=flex, at={($(thumb.south east)+(10pt,0)$)}, anchor=south west, xlabel={Finger flexion (rad)}, xmin=-0.1, xmax=1.6, xtick={0,0.5,1,1.5}]
\addplot[avline, black!45] table[col sep=comma, x=bin_center_rad, y=teleop]{figures/data/action_gap/a_hist_rhand_flexion_all12.csv};
\addplot[avline, gapOurs] table[col sep=comma, x=bin_center_rad, y=idm]{figures/data/action_gap/a_hist_rhand_flexion_all12.csv};
\addplot[avline, gapEgo] table[col sep=comma, x=bin_center_rad, y=retarget]{figures/data/action_gap/a_hist_rhand_flexion_all12.csv};
\end{axis}
\begin{axis}[avbase, name=grad, at={($(flex.south east)+(36pt,0)$)}, anchor=south west,
    width=0.22\textwidth, xmin=-0.55, xmax=1.55, ymin=-0.25, ymax=0.75,
    xtick={0,1}, xticklabels={{All\\params}, {Action\\expert}}, x tick label style={align=center},
    ytick={0,0.25,0.5,0.75}, ylabel={Gradient cosine}, ylabel style={yshift=-3pt}]
\draw[black!60] (axis cs:-0.55,0) -- (axis cs:1.55,0);
\addplot[ybar, bar width=0.22, bar shift=0pt, fill=gapOurs, draw=none] coordinates {(-0.3,0.532)};
\node[font=\tiny, anchor=south, inner sep=1pt] at (axis cs:-0.3,0.542) {0.53};
\addplot[ybar, bar width=0.22, bar shift=0pt, fill=gapEgo, draw=none] coordinates {(0.00,0.170)};
\node[font=\tiny, anchor=south, inner sep=1pt] at (axis cs:0.00,0.180) {0.17};
\addplot[ybar, bar width=0.22, bar shift=0pt, fill=gapEgo!45, draw=none] coordinates {(0.3,0.138)};
\node[font=\tiny, anchor=south, inner sep=1pt] at (axis cs:0.3,0.148) {0.14};
\addplot[ybar, bar width=0.22, bar shift=0pt, fill=gapOurs, draw=none] coordinates {(0.7,0.510)};
\node[font=\tiny, anchor=south, inner sep=1pt] at (axis cs:0.7,0.520) {0.51};
\addplot[ybar, bar width=0.22, bar shift=0pt, fill=gapEgo, draw=none] coordinates {(1.00,-0.033)};
\node[font=\tiny, anchor=north, inner sep=1pt] at (axis cs:1.00,-0.043) {$-$0.03};
\addplot[ybar, bar width=0.22, bar shift=0pt, fill=gapEgo!45, draw=none] coordinates {(1.3,-0.052)};
\node[font=\tiny, anchor=north, inner sep=1pt] at (axis cs:1.3,-0.135) {$-$0.05};
\end{axis}
\begin{axis}[avbase, name=roll, at={($(grad.south east)+(36pt,0)$)}, anchor=south west,
    width=0.16\textwidth, xmin=-0.6, xmax=1.6, ymin=-12, ymax=100,
    xtick={0,1}, xticklabels={{In-\\domain}, {Out-of-\\domain}}, x tick label style={align=center},
    ytick={0,50,100}, ylabel={Close to teleop.\ (\%)}, ylabel style={yshift=-3pt}]
\addplot[gapOurs, thick, mark=*, mark size=1.8pt] coordinates {(0,67.2) (1,54.9)};
\addplot[gapEgo, thick, mark=*, mark size=1.8pt] coordinates {(0,40.0) (1,7.7)};
\addplot[gapEgo!45, thick, mark=*, mark size=1.8pt] coordinates {(0,36.6) (1,10.0)};
\node[font=\tiny, gapOurs, anchor=east] at (axis cs:-0.1,67.2) {67};
\node[font=\tiny, gapEgo, anchor=east] at (axis cs:-0.1,47.0) {40};
\node[font=\tiny, gapEgo!70, anchor=east] at (axis cs:-0.1,29.0) {37};
\node[font=\tiny, gapOurs, anchor=west] at (axis cs:1.1,54.9) {55};
\node[font=\tiny, gapEgo!70, anchor=west] at (axis cs:1.1,19.0) {10};
\node[font=\tiny, gapEgo, anchor=west] at (axis cs:1.1,0.0) {8};
\end{axis}
\path (thumb.south) ++(0,-28pt) coordinate (lab);
\coordinate (amid) at ($(thumb.south east)!0.5!(flex.south west)$);
\node[font=\footnotesize, anchor=base] at (amid |- lab) {(a) Right-hand actions};
\node[font=\footnotesize, anchor=base] at (grad.south |- lab) {(b) Co-training with teleop.};
\node[font=\footnotesize, anchor=base] at (roll.south |- lab) {(c) Robot rollouts};
\end{tikzpicture}
\caption{\textbf{Action gap: distribution and co-training.} (a)~Right-hand actions over all tasks. (b)~Cosine between the policy gradients from teleoperated PnP data and from the out-of-domain tasks. (c)~Share of hand actions within the teleoperated distribution in real robot rollouts.}
\label{fig:action_vocab}
\vspace{-0.3in}
\end{figure}

We then examine how these labels affect policy learning by comparing the actions and the training gradients they induce (see \Cref{fig:action_vocab}). Retargeted hand actions differ substantially from teleoperated actions (see \Cref{fig:action_vocab}a). At the pretrained weights of $\pi_{0.5}$, we compute the gradient of the flow-matching loss on teleoperated in-domain data and on the data of each out-of-domain task. The cosine similarity between the two gradients measures how much the two sources agree on the update direction (\Cref{app:action_vocab}). Our data yields a cosine similarity of 0.53, compared with 0.17 and 0.14 for the baselines, and retargeted data yields slightly negative similarity within the action expert (see \Cref{fig:action_vocab}b). This suggests that retargeted labels steer the policy in a direction that conflicts with the teleoperated data, which may explain why the baselines gain little on out-of-domain tasks. Consistently, the baselines' hand actions during robot execution stay close to the teleoperated distribution on in-domain tasks but diverge on out-of-domain tasks, whereas our policy stays close on both (see \Cref{fig:action_vocab}c).

\Needspace*{18\baselineskip}
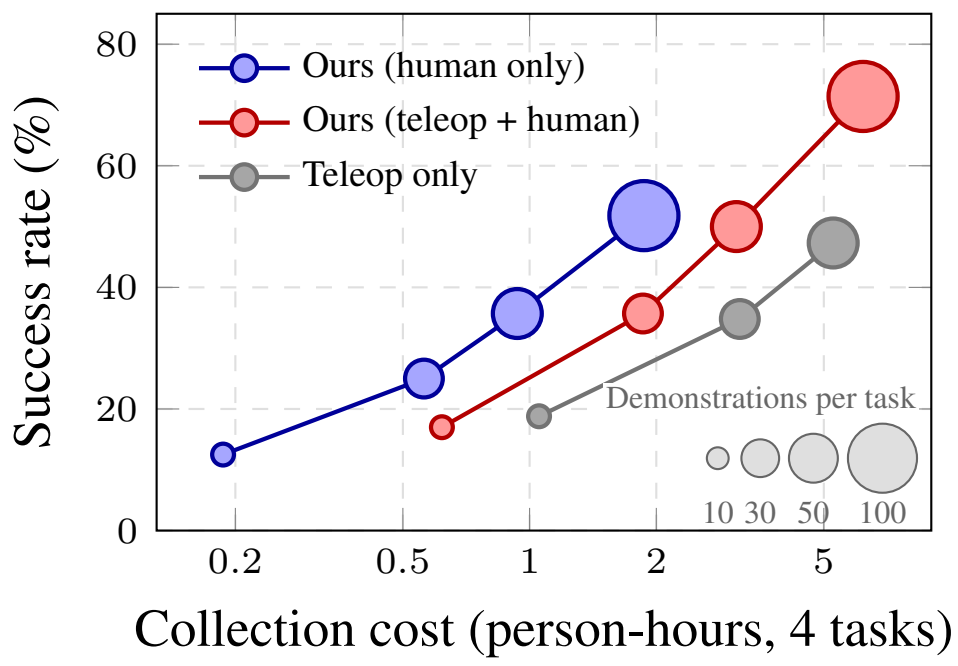
\begin{wrapfigure}[15]{r}{0.48\columnwidth}
\vspace{-\intextsep}
\centering
\begin{tikzpicture}
\begin{axis}[
    width=\linewidth, height=5.0cm,
    xmode=log, log ticks with fixed point,
    xtick={0.2,0.5,1,2,5},
    xmin=0.13, xmax=9,
    ymin=0, ymax=85, ytick={0,20,...,80},
    xlabel={Collection cost (person-hours, 4 tasks)},
    ylabel={Success rate (\%)},
    xlabel style={font=\scriptsize}, xlabel near ticks,
    ylabel style={font=\scriptsize, yshift=6pt}, ylabel near ticks,
    tick label style={font=\scriptsize},
    grid=major,
    grid style={dashed, gray!25},
    legend cell align=left,
    legend pos=north west,
    legend style={font=\scriptsize, draw=none, fill=none},
    clip=false,
]
\addplot[
    blue!60!black, thick, mark=none,
    visualization depends on={\thisrow{ms} \as \ms},
    nodes near coords={},
    nodes near coords style={anchor=center},
    every node near coord/.append style={
        circle, draw=blue!60!black, fill=blue!35,
        thick, inner sep=0pt, minimum size=\ms pt
    },
    legend image post style={
        mark=*, mark size=2.4pt, mark options={fill=blue!35}
    }
] table[x=t, y=sr] {
    t     sr    ms
    0.187 12.5  4.1
    0.561 25.0  7.1
    0.935 35.7  9.2
    1.869 51.8 13.0
};
\addlegendentry{Ours (human only)}
\addplot[
    red!70!black, thick, mark=none,
    visualization depends on={\thisrow{ms} \as \ms},
    nodes near coords={},
    nodes near coords style={anchor=center},
    every node near coord/.append style={
        circle, draw=red!70!black, fill=red!40,
        thick, inner sep=0pt, minimum size=\ms pt
    },
    legend image post style={
        mark=*, mark size=2.4pt, mark options={fill=red!40}
    }
] table[x=t, y=sr] {
    t     sr    ms
    0.619 17.0  4.1
    1.859 35.7  7.1
    3.099 50.0  9.2
    6.198 71.4 13.0
};
\addlegendentry{Ours (teleop + human)}
\addplot[
    gray!90!black, thick, mark=none,
    visualization depends on={\thisrow{ms} \as \ms},
    nodes near coords={},
    nodes near coords style={anchor=center},
    every node near coord/.append style={
        circle, draw=gray!90!black, fill=gray!70,
        thick, inner sep=0pt, minimum size=\ms pt
    },
    legend image post style={
        mark=*, mark size=2.4pt, mark options={fill=gray!70}
    }
] table[x=t, y=sr] {
    t     sr    ms
    1.053 18.8  4.1
    3.158 34.8  7.1
    5.263 47.3  9.2
};
\addlegendentry{Teleop only}

\coordinate (leg) at (rel axis cs:0.985,0.14);
\begin{scope}[black!60, font=\tiny]
    \node[anchor=south east, inner sep=0.5pt, fill=white]
        at ([yshift=7.5pt]leg) {Demonstrations per task};
    \draw[fill=gray!25] ([xshift=-7pt]leg)  circle [radius=6.5pt];
    \draw[fill=gray!25] ([xshift=-20pt]leg) circle [radius=4.6pt];
    \draw[fill=gray!25] ([xshift=-30pt]leg) circle [radius=3.55pt];
    \draw[fill=gray!25] ([xshift=-38pt]leg) circle [radius=2.05pt];
    \node[anchor=north, inner sep=0pt]
        at ([xshift=-7pt,yshift=-8pt]leg) {100};
    \node[anchor=north, inner sep=0pt]
        at ([xshift=-20pt,yshift=-8pt]leg) {50};
    \node[anchor=north, inner sep=0pt]
        at ([xshift=-30pt,yshift=-8pt]leg) {30};
    \node[anchor=north, inner sep=0pt]
        at ([xshift=-38pt,yshift=-8pt]leg) {10};
\end{scope}
\end{axis}
\end{tikzpicture}
\vspace{-6pt}
\caption{\textbf{Collection cost.} Success rate versus person-hours on four tasks. Marker size indicates demonstrations per task.}
\label{fig:cost}
\end{wrapfigure}
\textbf{Scaling and cost.}
\label{sec:scaling}
To assess whether human videos are a cost-effective source of demonstrations, we vary the number of demonstrations per task on the four in-domain tasks and compare the success rate against collection cost in person-hours (see \Cref{fig:cost}). Recording a human video takes about a fifth of the time of a teleoperated episode (see \Cref{tab:cost}). Notably, training with translated human videos alone matches the success rate obtained with 50 teleoperated demonstrations per task at about one third of the collection cost. This suggests that, although a translated human video yields a slightly lower success rate per demonstration than a teleoperated episode, it is more cost-effective in collection time. Moreover, augmenting the teleoperated dataset with an equal number of human videos increases the success rate from 47.3\% to 71.4\% and requires less than one additional hour of recording.

\begin{wraptable}[12]{r}{0.54\textwidth}
\vspace{-11pt}
\centering
\caption{\textbf{Execution speed.} Success rate (\%) on the OpenArm and mean time from rollout start to object placement on the four PnP tasks.}
\label{tab:speed}
\footnotesize
\setlength{\tabcolsep}{3pt}
\begin{tabular}{@{}lcccc@{}}
\toprule
Method & Speed & In-dom. & Out-of-dom. & Time (s) \\
\midrule
Teleop only & $1\times$ & 47.3 & \phantom{0}3.1 & 14.2 \\
\midrule
\textbf{Ours} (as recorded) & $1.25\times$ & 75.9 & \textbf{73.0} & 11.9 \\
\midrule
\multirow{2}{*}{\textbf{Ours} (retimed)} & $1\times$ & \textbf{85.7} & 70.4 & 13.8 \\
 & $2\times$ & 57.1 & 42.9 & \textbf{\phantom{0}7.4} \\
\bottomrule
\end{tabular}
\vspace{-0.1in}
\end{wraptable}
\textbf{Execution speed.}
Since human demonstrators move at roughly $1.25\times$ the speed of the teleoperated demonstrations, we examine how the speed of synthesized demonstrations affects the policy. Using the same keyframes, we retime the skeleton flow to the teleoperated speed ($1\times$) or twice that speed ($2\times$) before synthesis and compare with videos synthesized at the recorded speed (see \Cref{tab:speed}). Matching the teleoperated speed improves the in-domain success rate from 75.9\% to 85.7\% while keeping the out-of-domain success rate similar. This suggests that the policy inherits the speed of its training demonstrations, as its task completion time follows the speed of the synthesized videos. However, training at $2\times$ produces the fastest policy at the cost of a lower success rate on both task groups.

\subsection{Ablations}
\label{sec:ablations}

We ablate the components of \ours and the policy backbone to verify that each component contributes to policy success and that the gains do not depend on a specific VLA. All ablations use the OpenArm, and per-task results are given in \Cref{tab:ablation_full,tab:backbone_full}.

\textbf{Pipeline components.}
As shown in \Cref{tab:ablation}, removing any of the four components reduces the success rate on both task groups. Without the autoregressive reference, the robot appearance stays consistent but the scene changes across keyframes. Without subsequent keyframes, the video model is conditioned only on the first keyframe, so the motion is preserved but the objects and the scene deviate from the human video. Without the hand skeleton flow, the video model receives no motion guidance between keyframes, and the synthesized robot no longer follows the demonstrator, which causes the largest drop. This shows that each stage is needed to keep the synthesized demonstration faithful to the human video in scene, appearance, and motion. Interestingly, removing the wrist correction reduces the success rate more on out-of-domain tasks, since the policy relies more on generated action labels for tasks without teleoperated demonstrations, and many of their motions lie outside the training distribution of the IDM. We show failure cases of each variant in \Cref{app:ablations}.

\textbf{Policy backbone.}
To verify that the benefit of \ours is not specific to $\pi_{0.5}$, we also fine-tune RLDX-1-Img~\citep{rldx2026}, a VLA with a different architecture, on the same data (see \Cref{tab:backbone}). \ours consistently improves the success rate of RLDX-1-Img, from 53.6\% to 60.7\% on in-domain tasks and from 0.0\% to 40.3\% on out-of-domain tasks. This demonstrates that the demonstrations synthesized by \ours remain effective across policy architectures.

\begin{table*}[t]
\centering
\caption{\textbf{Ablation study of \ours on OpenArm.} Success rates (\%) are averaged over four in-domain and seven out-of-domain tasks.}
\vspace{0.1in}
\label{tab:ablation_results}

\begin{subtable}[t]{0.49\textwidth}
    \centering
    \captionsetup{justification=centering}
    \caption{Component ablation.}
    \label{tab:ablation}
    \footnotesize
    \setlength{\tabcolsep}{1.5pt}
    \begin{tabular}{@{}lcc@{}}
        \toprule
        Method & In-dom. & Out-of-dom. \\
        \midrule
        \oursb & \textbf{75.9} & \textbf{73.0} \\
        \quad w/o Autoregressive reference & 68.8 & 66.3 \\
        \quad w/o Subsequent keyframes & 42.9 & 39.3 \\
        \quad w/o Hand skeleton flow & 42.0 & 21.4 \\
        \quad w/o IDM wrist correction & 65.2 & 48.0 \\
        \bottomrule
    \end{tabular}
\end{subtable}
\hfill
\begin{subtable}[t]{0.49\textwidth}
    \centering
    \vspace{0.04in}
    \captionsetup{justification=centering}
    \caption{Policy backbone comparison.}
    \label{tab:backbone}
    \footnotesize
    \setlength{\tabcolsep}{5pt}
    \begin{tabular}{@{}lcc@{}}
        \toprule
        Method & In-dom. & Out-of-dom. \\
        \midrule
        $\pi_{0.5}$ (Teleop only) & 47.3 & 3.1 \\
        \quad + \oursb & \textbf{75.9} & \textbf{73.0} \\
        \midrule
        RLDX-1-Img (Teleop only) & 53.6 & 0.0 \\
        \quad + \oursb & \textbf{60.7} & \textbf{40.3} \\
        \bottomrule
    \end{tabular}
\end{subtable}
\end{table*}

\section{Conclusion}
\label{sec:conclusion}
We presented \ours, which turns a human video into a robot demonstration
the way virtual try-on changes the garment on a person: it keeps the
demonstrated motion and swaps the embodiment. \ours translates
human keyframes into robot keyframes, synthesizes the robot video
between them under the guidance of the hand skeleton, and labels its
actions with an inverse dynamics model. On two real robots with
dexterous hands, the synthesized demonstrations teach a policy seven
tasks without any robot demonstration, including bimanual ones. Human
video translated this way also matches the success rate of teleoperated demonstrations at
about a third of their collection cost. We discuss limitations and future
work in \Cref{app:limitation}.

\section*{Acknowledgement}
This paper was supported by RLWRLD Inc. We thank Beomjun Kim, Junhyeong Park, Jungwoo Park, and Seunghoon Sim for their help in collecting robot and human demonstrations for our real-world experiments.

\vspace{0.2in}

\bibliographystyle{plainnat}
\bibliography{references}

\appendix
\clearpage

\crefalias{section}{appendix}
\crefalias{subsection}{subappendix}
\crefalias{subsubsection}{subsubappendix}

\startcontents[appendix]
\section*{Appendix Contents}
\printcontents[appendix]{}{1}{\setcounter{tocdepth}{2}}
\clearpage
\newpage
\appendix

\section{Limitations and Future Work}
\label{app:limitation}

\textbf{Limitations.}
Adding an embodiment still requires a few hours of teleoperated video of
that robot, on any tasks, to teach the generative models and the inverse
dynamics model what it looks like and how it moves. This cost is paid
once per robot rather than once per task, but it is not zero.
The actions are pseudo labels: the IDM reads the joints from pixels, and
the hand reconstruction provides the wrist position. They cannot capture
forces, and their precision is bounded by what stereo video and a hand
reconstruction resolve, so tasks that depend on forces or on fine
precision, such as tight insertion, are not yet labeled accurately
enough. The wrist correction also relies on the head camera calibration.
The human videos are recorded from the position of the robot's head
camera, and the teleoperated data keep the neck and the base of the
robot fixed. Under this fixed viewpoint, \ours translates human videos
into each embodiment reliably (see \Cref{fig:embodiments}). With a
moving camera, as in in-the-wild egocentric videos, the input falls
outside the training distribution of the video model and the IDM. In
particular, the IDM always predicts fixed neck and base actions, since
the neck and the base never move in its training data.
The demonstrated motion must be one the robot can execute. The pipeline
can retime the human video to the robot's pace, but it cannot correct a
motion toward a position beyond the reach of the robot arm or a grasp
that the robot hand cannot form.
Our evaluation covers tabletop manipulation on two robots. We also
translate human videos into the GR-1, but could not evaluate policies on
it without access to the physical robot.
Finally, synthesis trades collection time for compute
(see~\Cref{tab:cost}), which favors tasks that are expensive to
teleoperate.

\textbf{Future work.}
The natural next step is scale: translating large egocentric corpora
such as EgoDex into every embodiment we own, so that a robot's training
data grows with the supply of human video rather than with the supply
of operators.
Since the policy learns from pseudo labels, improving the accuracy of
the IDM is another key direction. Policy rollouts on the robot are
recorded with their executed actions, so they naturally become training
data for the IDM, and the IDM can keep improving as the policy is
deployed.
Currently, the IDM and the hand reconstruction estimate actions
separately, and we simply replace the wrist position predicted by the
IDM with the reconstructed one. An IDM that takes the reconstructed
wrist position as an additional input could combine the two estimates
and predict more accurate actions.

\section{Implementation Details}
\label{app:impl}

\subsection{Policy Training Details}
\label{app:policy}

$\pi_{0.5}$ is a mixture of transformers in which the action expert
shares attention with the vision-language model, whereas RLDX-1-Img
follows the design of GR00T N1 \citep{nvidia2025gr00t}, in which a
separate action head is conditioned on vision-language features.
Both $\pi_{0.5}$ and RLDX-1-Img are fully fine-tuned for 30k steps
with a batch size of 64 on four H100 GPUs, with the same settings for
every data mixture of \Cref{tab:main_real}. $\pi_{0.5}$ is
trained with AdamW ($\beta_1 = 0.9$, $\beta_2 = 0.95$,
$\epsilon = 10^{-8}$, weight decay 0.01) at a learning rate of
$2.5 \times 10^{-5}$ with cosine decay to $2.5 \times 10^{-6}$,
gradient clipping at norm 1.0, and an action chunk of 50 steps (2.5\,s).
States and actions are normalized by their 1st and 99th percentiles to
$[-1, 1]$. RLDX-1-Img follows the configuration of its official
repository: the vision-language backbone is frozen except for its top
four language layers, the action expert is trained in full (2.4B of
6.9B parameters in total), the learning rate is $10^{-4}$ with cosine
decay, and the action chunk is 16 steps (0.8\,s), and the optimizer and
the normalization are the same as for $\pi_{0.5}$.

\subsection{RoboTryOn Pipeline Details}
\label{app:robotryon}

\subsubsection{Keyframe Selection}
\label{app:keyframe}
Keyframes are selected from the 3D wrist trajectories of the
reconstructed hands (\Cref{app:skeleton}), smoothed in time with a
Gaussian of $\sigma = 0.25$\,s. A frame is a candidate when the larger of
the two wrist speeds has a local minimum within $\pm 1$\,s: the hands
are unblurred there, and a hand has typically arrived at a new location.
The next keyframe is the first candidate at which a wrist has moved more
than a threshold from its position in the previous keyframe. The
threshold is set for each episode as the 30th percentile of the wrist
displacements between all pairs of up to 400 frames sampled uniformly
from the episode, where the displacement of a pair is the larger 3D
movement of the wrists visible in both frames. Because the
reconstruction is most reliable when a hand is slow and fully visible,
this rule also tends to avoid frames in which a hand is undetected or
detected with low confidence. The base video model supports clips of up
to 20 seconds, but for the consistency of the synthesized video we cap
the interval between consecutive keyframes at 145 frames, about 7.25
seconds at 20~Hz. When no candidate satisfies the rule within the cap,
we take the frame with the largest displacement among those whose
reconstruction confidence is at least the median of the interval, or the
frame at the cap if no frame has a confidence. The first and last frames
are always keyframes. Pick-and-place videos yield about four keyframes
and the longer tasks six to seven.

\subsubsection{Hand Skeleton}
\label{app:skeleton}

\textbf{Hand reconstruction.}
The human hands are reconstructed with HaWoR \citep{zhang2025hawor}
from each view of the stereo camera. The depth of the wrist is corrected
by triangulating the two views, and the left and right hand predictions
of the two views complement each other, so that a hand that leaves the
frame in one view and is misdetected there is recovered from the other
view.

\textbf{Keypoint set.}
Each hand is represented by 17 keypoints: four on the thumb, three on
each of the other four fingers (the knuckle, the middle joint, and the
tip), and the wrist. The skeleton connects the keypoints of each finger
in sequence and connects the finger bases to one another and to the
wrist. The same set is defined for
the human hand, the Inspire RH56F1, the WujiHand2, and the GR-1 hand, so
that hands with different degrees of freedom map to the same keypoints.
For a robot the keypoints come from forward kinematics of the recorded
joint state, projected into each camera view, and for a person they come
from the reconstructed hand, retargeted to the target robot hand.
Because the retargeting is imprecise, keypoints are perturbed and
randomly dropped during training: each keypoint is displaced by up to
half the length of the finger segment it belongs to, the whole skeleton
is dropped with probability 0.2, and each keypoint is dropped
independently with probability 0.2. The model thus learns to treat the
skeleton as a coarse guide while its prior over robot hands, learned
from the robot data, determines the configuration the generated hand
takes. The retargeting error does not enter the action labels, which
are read from the generated video by the IDM. The skeleton is rendered
as its 2D projection, drawn over the frame for the image model and as
a video of the same length for the video model.

\subsubsection{Keyframe Translation}
\label{app:mask}

\textbf{Mask construction.}
The mask is shape-agnostic in the sense of virtual try-on: it is
enlarged and smoothed so that the model cannot recover the original
silhouette of the arm and hand it covers. For a robot frame the mask
is built from a rendering of the robot from its URDF at the recorded
joint state and a SAM3 segmentation of the robot, which are
complementary, since neither alone covers the robot exactly. During
training the mask is further deformed to random sizes and shapes so
that the model is robust to its size and outline, and, with some
probability and where the object is known, the region of the
manipulated object is removed from the mask. At inference the mask of
the human arm and hand comes from SAM3 alone, dilated so that the shape
cannot be inferred, and the region of the manipulated object is always
removed so that the object is preserved.

\textbf{Reference sampling.}
The reference frame is sampled from the same episode two to five
seconds before the target, since a closer frame would allow the model to copy
the reference instead of learning to move the robot.

\textbf{Stereo generation.}
Both views are generated as one token sequence with an additional rotary
position dimension separating left and right tokens, and training mixes
single-view and stereo examples.

\textbf{Autoregressive inference.}
The denoised latent of $\hat\rvo^r_{t_{j-1}}$ is reused directly as the
reference tokens for $\hat\rvo^r_{t_j}$, without decoding and re-encoding.
\label{app:firstframe}

\textbf{First keyframe.}
The first keyframe has no previously translated keyframe to serve as
reference, so its reference is a real frame of the target robot with
everything but the robot masked out, which selects the embodiment. To
train for this case, the reference is replaced by such a frame with
probability 0.05 during training.

\subsubsection{Video Synthesis}
\label{app:video}
All videos and actions are recorded, synthesized, and labeled at 20~Hz.
A demonstration is cut at keyframe times into clips of at most 145
frames (\Cref{app:keyframe}). Each clip is synthesized with the
two to four keyframes that fall inside it, adjacent clips share the
keyframe at their boundary, and the clips are concatenated in time. The
rendered skeleton flow is encoded by the video VAE like the clip itself
and attended to as in-context tokens.
Demonstrations of 10--25 seconds are synthesized in this way
(\Cref{fig:longvideo}).
\subsubsection{Retiming}
By default the human video is synthesized as recorded. For
\Cref{tab:speed} each human video is divided by its action into
phases, approaching the object, grasping it, lifting it, and placing it,
into which the tasks without robot demonstrations divide as well. For
each phase the skeleton flow is resampled in time so that its speed
matches the mean action speed of the teleoperated episodes in that
phase, at the chosen multiple, and the video is synthesized again from
the same keyframes at their new temporal positions. Human demonstrators
grasp and lift faster than the teleoperators in many episodes, so these
are the phases the retiming changes most.

\subsubsection{Inverse Dynamics Model}
\label{app:idm}
The IDM takes frames $t$ and $t{+}20$ (1\,s at 20\,Hz) at
224$\times$224, stereo for the OpenArm and RB-Y1 and monocular for the
GR-1, and predicts a 20-step chunk of the full joint vector (28, 66,
and 44 dimensions). Following the DreamGen IDM \citep{jang2025dreamgen},
the three embodiments share one model and are distinguished by an
embodiment tag, and the action head is a flow-matching head on the
Depth Anything~V2 backbone. Of the 24 layers of its ViT, the first 11
are frozen and the rest are fine-tuned, and the Depth Anything head is
not used. For labeling, the model is applied in a
sliding window, overlapping predictions are averaged, and the sequence
is smoothed with a Savitzky--Golay filter \citep{savitzky1964smoothing}
of polynomial order 2 over a window of 9 frames for the arm joints and
5 frames for the hand joints. The states
$\hat\rvq_t$ of a synthesized demonstration are derived from its action
labels by the relation measured on the teleoperated data, where the
recorded state lags the commanded action by four steps, so that
$\hat\rvq_t = \hat\rva_{t-4}$.
\subsubsection{End-Effector Correction}
The human wrist position is taken from the hand reconstruction of
\Cref{app:skeleton}, in the camera frame, and transformed into
the robot base frame with the head camera calibration, since the human
video is recorded from the position of the robot's head camera. The
robot hand implied by the IDM's joints is scaled, finger segment by
finger segment, to the size of the reconstructed human hand, and the
target wrist position is the one at which the end effector of the
scaled hand coincides with the reconstructed wrist. Since the
reconstruction is noisy, the target is taken only from segments with
high reconstruction confidence and smoothed in time, so that the
correction moves the arm gradually rather than snapping it to the
reconstruction. The target orientation is the wrist orientation
obtained by forward kinematics from the IDM's arm joints, so that only
the translation is corrected. The arm joints are then solved by inverse
kinematics in MuJoCo, initialized from the IDM's arm joints of the same
frame, and the hand joints are left unchanged. The correction is applied to
both arms for bimanual demonstrations.
\Cref{sec:ablations} ablates it.

\subsubsection{Training Setup}
\label{app:trsetup}
\begin{table}[t]
\centering
\footnotesize
\caption{\textbf{Training configuration.} Resolutions are per view.}
\label{tab:train}
\setlength{\tabcolsep}{4pt}
\begin{adjustbox}{max width=\textwidth}
\begin{tabular}{l ccc}
\toprule
& Image model $f_\theta$ & Video model $g_\phi$ & IDM $h_\psi$ \\
\midrule
Base model          & FLUX.2 [klein] 9B Base & LTX-2.5 & Depth Anything V2-L \\
Trained parameters  & full & LoRA (rank 64) & ViT (layers 1--11 frozen) + action head \\
Input resolution    & 1024$\times$576 & 512$\times$288, then 1024$\times$576 & 224$\times$224 \\
Views               & stereo : mono = 3 : 1 & stereo + mono & stereo (GR-1: mono) \\
Clip length / chunk & -- & $\le$145 frames @ 20 Hz & 20 frames @ 20 Hz \\
Steps               & 5k & 4k + 1k & 60k \\
Global batch size   & 64 & 16 & 128 \\
Learning rate       & $10^{-5}$, cosine & $10^{-4}$ & $10^{-4}$, cosine \\
Optimizer           & AdamW (0.9, 0.999) & AdamW (0.9, 0.999) & AdamW (0.9, 0.999) \\
Hardware            & 4$\times$H100 & 4$\times$H100, then 4$\times$H200 & 4$\times$H200 \\
\bottomrule
\end{tabular}
\end{adjustbox}
\end{table}

\textbf{Training corpus.}
\label{app:data}
All three models, the image model, the video model, and the IDM, are
trained on the same corpus of teleoperated robot episodes: 3.79 hours
(453 episodes) collected on the OpenArm with the Inspire RH56F1 hand,
2.53 hours (352 episodes) on the RB-Y1 with the WujiHand2, and 2.50 hours
(1,180 clips) of the GR-1 taken from the public GR00T teleoperation
dataset released with DreamDojo \citep{gao2026dreamdojo}. The OpenArm and
RB-Y1 recordings are stereo, whereas the GR-1 recordings are monocular, so the
generative models learn stereo generation from the first two. The corpus
consists of the teleoperated demonstrations of the four pick-and-place
tasks used in \Cref{sec:experiments} plus additional teleoperated
recordings. The additional recordings are used only to train the three
models above and play no part in training any VLA policy.

\textbf{Training configuration.}
\label{app:train}
The image model $f_\theta$ is a fully fine-tuned FLUX.2 [klein] 9B Base
\citep{bfl2026fluxklein}, the video model $g_\phi$ an in-context LoRA
on LTX-2.5 \citep{lightricks2026ltx25,hacohen2026ltx2}, and the IDM
$h_\psi$ a Depth Anything~V2 backbone \citep{yang2024depthv2} with the
action head of the DreamGen IDM \citep{jang2025dreamgen}.
\Cref{tab:train} lists their training configuration, and all three are
trained on the corpus of \Cref{app:data}.
The image model is trained with AdamW ($\beta_1 = 0.9$, $\beta_2 = 0.999$) and a cosine learning rate
schedule. Its input embedder does not accept the channel-wise
concatenation of the conditioned frame with the noisy target, so the
embedder is extended by zero-initialized input channels that are
trained with the rest of the model. Training examples are stereo and
single-view in a 3:1 ratio, and the GR-1 recordings are single-view only.
The video model LoRA has rank 64 and is trained with AdamW ($\beta_1 = 0.9$, $\beta_2 = 0.999$) in two stages matching
the two-stage inference of \Cref{app:infer}: 4k steps at
512$\times$288 per view followed by 1k steps at 1024$\times$576 per
view.

\subsection{Baselines}
\label{app:base}

\begin{table}[t]
\centering
\small
\caption{\textbf{Inference configuration and cost per unit.} Time is
measured on a single H100. Video stage 2
upsamples and refines the output of stage 1.}
\label{tab:infer}
\setlength{\tabcolsep}{4pt}
\begin{tabular}{l l c c c c}
\toprule
Stage & Model & Output per view & Views & Steps & Time (s) \\
\midrule
Hand reconstruction   & HaWoR                 & 3D hand pose, per frame & 2 & --  & 0.448 / frame \\
Keyframe translation  & $f_\theta$            & 1024$\times$576         & 2 & 30 & 34.6 / keyframe \\
Video synthesis, stage 1 & $g_\phi$           & 512$\times$288         & 2 & 8   & 50.0 / clip \\
Video synthesis, stage 2 & $g_\phi$            & 1024$\times$576         & 2 & 3   & 50.0 / clip \\
Action labeling       & $h_\psi$              & 20-step joint chunk     & 2 & -- & 0.077 / window \\
End-effector correction & MuJoCo IK           & arm joints, per frame   & -- & -- & 0.22 / frame \\
\bottomrule
\end{tabular}
\end{table}

\begin{table}[t]
\centering
\small
\caption{\textbf{Pipeline wall-clock time} for a 15\,s demonstration
(300 frames, one embodiment) on a single H100.}
\label{tab:pipeline}
\setlength{\tabcolsep}{6pt}
\begin{tabular}{l c c}
\toprule
Stage & Units & Time (s) \\
\midrule
Hand reconstruction + keyframe selection & 300 frames & 134 \\
Keyframe translation                      & 5 keyframes & 173 \\
Video synthesis, stage 1                  & 3 clips & 150 \\
Video synthesis, stage 2                  & 3 clips & 150 \\
Action labeling                           & 281 windows & 22 \\
End-effector correction                   & 300 frames & 66 \\
\bottomrule
\end{tabular}
\end{table}

\textbf{Common setup.}
\label{app:baselines}
Both baselines are re-implemented on our hardware with the same
backbone, teleoperated episodes, and human videos as \ours.

\textbf{Retargeting.}
Both use the same pseudo actions: the 3D hand pose reconstructed with
HaWoR is mapped to the robot hand by a geometric mapping of joint angles
(finger flexion angles matched to the hand's joint ranges, wrist pose
transformed into the base frame with the head camera calibration)
following AnyTeleop \citep{qin2023anyteleop}, and the arm joints are
obtained by inverse kinematics in MuJoCo from the wrist pose.

\textbf{EgoMimic.}
The human arm and hand are masked in every human frame. In the robot
episodes the robot arm is masked from the recorded state, by
projecting the URDF into the camera. At test time the mask is placed
on the live image from the current robot state and the camera
calibration. The masking follows the official implementation.

\textbf{Masquerade.}
The human arm is inpainted out and a rendering of the robot from its
URDF in the retargeted state is composited onto the frame. Robot
episodes receive the same rendering from the recorded state, and at
test time the rendering is composited onto the live image from the
current state. The original uses E2FGVI \citep{li2022e2fgvi} for the
inpainting. We use ProPainter \citep{zhou2023propainter}, which leaves
fewer artifacts on our videos. Following the original, the policy is
trained with an auxiliary loss that predicts future 2D robot hand
keypoints from the image features, with targets from the retargeted
(human) or recorded (robot) state. The head, the prediction horizon,
and the loss weight follow the official implementation.
The original single-arm gripper pipeline is extended to both arms and
to the dexterous hands of our robots.

\subsection{Inference Details}
\label{app:infer_cost}

\textbf{Inference configuration and compute.}
\label{app:infer}
\Cref{tab:infer} lists the inference configuration of every stage
and its cost on a single H100. The video model runs in two stages: the
first generates the clip at 512$\times$288 per view with a distilled
model in 8 sampling steps, and the second upsamples the result to
1024$\times$576 per view and refines it in 3 steps. Both stages are
conditioned on the same keyframes and skeleton flow, the second at the
higher resolution, and both generate the two stereo views in one
sequence. The image model generates each keyframe at 1024$\times$576
per view in 30 sampling steps with a guidance scale of 3.5, and the
keyframes of a demonstration are generated sequentially since each
conditions on the previous one. The IDM predicts a 20-step action chunk
from each 1\,s window and is applied in a sliding window with a
stride of one frame over the demonstration.

\Cref{tab:pipeline} breaks down the wall-clock time of the full
pipeline for a typical 15\,s demonstration (300 frames at 20\,Hz,
translated into one embodiment). \Cref{tab:cost} accounts for collection in person-hours and \Cref{app:limitation} notes the trade of collection time for compute.

\section{Experimental Details}
\label{app:exper}

\subsection{Real-World Environment}
\label{app:env}

\begin{figure}[t]
\centering
\newcommand{\stp}[3]{%
  \begin{minipage}[t]{0.192\textwidth}
  \centering
  \thumbat{\linewidth}{settings/#1}{#2}\\[2pt]
  {\footnotesize #3}
  \end{minipage}}
\stp{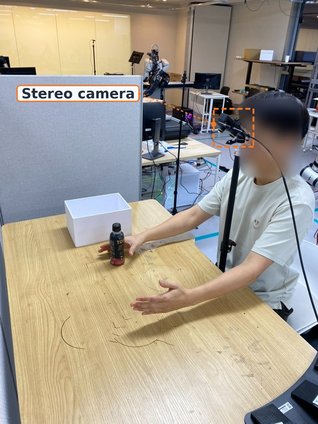}{human}{(a) Human video collection}\hfill
\stp{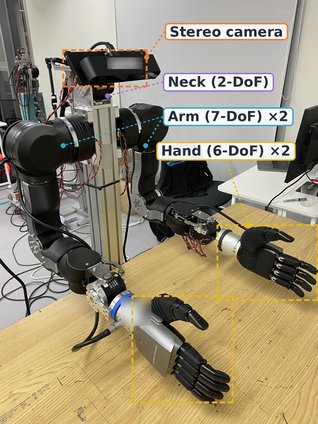}{openarm}{(b) OpenArm + RH56F1}\hfill
\stp{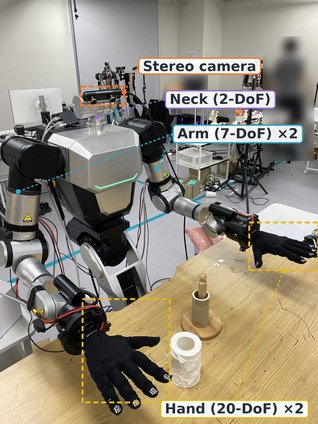}{rby1}{(c) RB-Y1 + WujiHand2}\hfill
\stp{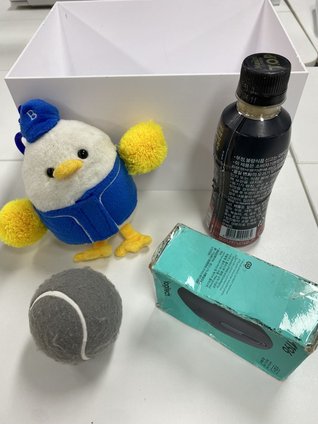}{objects_id}{(d) In-domain objects}\hfill
\stp{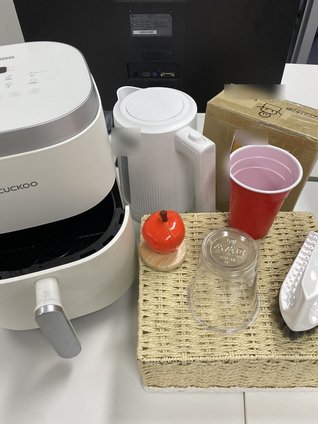}{objects_ood}{(e) Out-of-domain objects}
\caption{\textbf{Real-world setup.} (a)~Human video collection from the robot's camera position. (b)~OpenArm embodiment. (c)~RB-Y1 embodiment. (d)~Objects of the in-domain tasks. (e)~Objects introduced for the out-of-domain tasks.}
\label{fig:setup}
\end{figure}

\textbf{Hardware setup.}
We use two humanoid robots, OpenArm and RB-Y1 (\Cref{fig:setup}b,c), in our real-world experiments. Both humanoids are equipped with two 7-DoF arms. Each arm of the OpenArm is equipped with a 6-DoF Inspire RH56F1 hand, and each arm of the RB-Y1 is equipped with a 20-DoF WujiHand2 hand. The head of each robot is mounted on a 2-DoF neck that supports pitch and yaw motions. A stereo camera is mounted on each neck, and its left and right egocentric views are used as visual inputs throughout all experiments.

\begin{figure}[t]
\centering
\includegraphics[width=\textwidth]{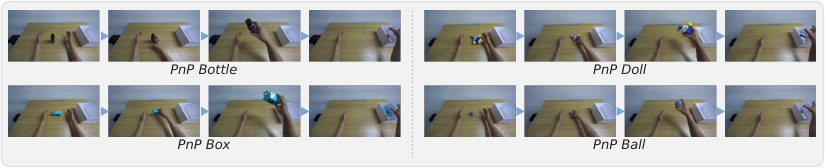}\\[1pt]
{\footnotesize (a) In-domain tasks}\\[4pt]
\includegraphics[width=\textwidth]{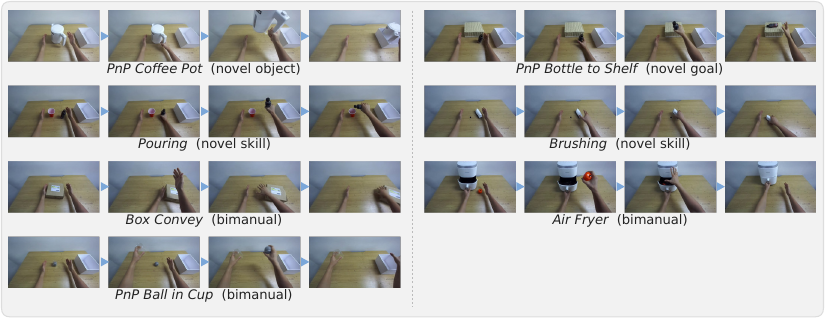}\\[1pt]
{\footnotesize (b) Out-of-domain tasks}
\caption{\textbf{Real-world tasks.} (a)~In-domain tasks with teleoperated demonstrations and human videos. (b)~Out-of-domain tasks with human videos only.}
\label{fig:task_details}
\end{figure}

\textbf{Task details.}
All tasks are formulated as language-conditioned manipulation tasks. We evaluate eleven tasks in total (\Cref{fig:tasks}), comprising four in-domain pick-and-place (PnP) tasks and seven out-of-domain tasks. The in-domain tasks (\Cref{fig:tasks}a) consist of \textit{PnP Bottle}, \textit{PnP Doll}, \textit{PnP Box}, and \textit{PnP Ball}, with 50 teleoperated demonstrations per robot embodiment and 50 human videos for each task. The human videos are recorded from the position of the robot's head camera (\Cref{fig:setup}a). The seven out-of-domain tasks (\Cref{fig:tasks}b) have 50 human videos each and no robot demonstrations: \textit{PnP Coffee Pot}, \textit{PnP Bottle to Shelf}, \textit{Pouring}, \textit{Brushing}, \textit{Box Convey}, \textit{Air Fryer}, and \textit{PnP Ball in Cup}.

The four in-domain PnP tasks share the same task structure, where the robot grasps a designated object and places it into a white container (\Cref{fig:setup}d). \textit{PnP Coffee Pot} follows the same pick-and-place structure with a novel coffee pot object. \textit{PnP Bottle to Shelf} requires the robot to grasp a bottle and place it at a designated location on a shelf. \textit{Pouring} requires the robot to grasp a bottle and pour its contents into a red cup. \textit{Brushing} requires the robot to grasp a brush and wipe the area marked with black tape. \textit{Box Convey} requires the robot to move a box to the right using both hands sequentially, with the left hand moving it first and the right hand moving it further. \textit{Air Fryer} requires the robot to open an air fryer, grasp an apple-shaped object, place it inside, and close the door. Finally, \textit{PnP Ball in Cup} requires the robot to lift an inverted cup with the left hand, grasp the ball with the right hand, and place it into the target container. \Cref{fig:setup}e shows the objects introduced by the out-of-domain tasks.

\textbf{Evaluation protocol.}
\label{app:eval}

We conduct 28 evaluation trials for each task. For single-arm tasks, a trial receives a score of 1 only when the task is successfully completed. For bimanual tasks, each hand contributes 0.5 to the score, yielding a maximum score of 1 when both required hand actions are successfully completed. The initial positions of the target objects are randomized within the workspace for each trial. For the PnP tasks, including \textit{PnP Bottle}, \textit{PnP Doll}, \textit{PnP Box}, \textit{PnP Ball}, \textit{PnP Coffee Pot}, and \textit{PnP Bottle to Shelf}, a trial is considered successful when the object is placed at the designated goal location. For \textit{Pouring}, the trial is successful when the bottle is tilted over the target cup. For \textit{Brushing}, success is determined by whether the brush reaches the designated region marked with black tape. For \textit{Box Convey}, the trial is successful when the specified sequential manipulation with the left and right hands is correctly executed. For \textit{Air Fryer}, success requires opening the air fryer, placing the apple-shaped object inside, and closing the door. For \textit{PnP Ball in Cup}, success requires lifting the cup with the left hand, grasping the ball with the right hand, and placing the ball at the designated goal location.

\subsection{Simulation Environment}
\label{app:sim_env}

\begin{figure}[t]
\centering
\includegraphics[width=\textwidth]{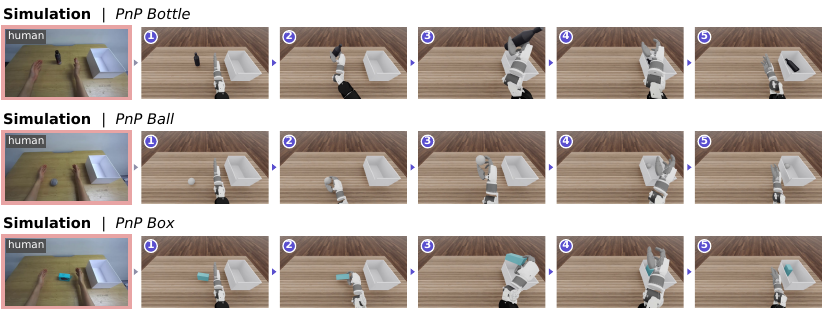}
\caption{\textbf{Simulated tasks.} Each row shows the human video of a
task (left) and a successful episode in simulation (1--5): the right
hand grasps the object, lifts it, and places it in the container.}
\label{fig:sim_tasks}
\end{figure}

\textbf{Setup.}
We build a simulated counterpart of the OpenArm setup in Isaac
Lab~\citep{mittal2023isaaclab}. The simulated robot follows the real
OpenArm: two 7-DoF arms with 6-DoF Inspire RH56F1 hands and a 2-DoF neck,
fixed at the downward pitch measured on the real robot. The head camera is positioned to match the mounting configuration of the real stereo camera, ensuring that the simulated egocentric view shares the same viewpoint as the real one (\Cref{fig:sim_tasks}).

\textbf{Tasks.}
We simulate the three in-domain pick-and-place tasks that have simulation
assets, \textit{PnP Bottle}, \textit{PnP Ball}, and \textit{PnP Box}.
\textit{PnP Doll} is excluded because no simulation asset of the doll
exists. For each human video, the objects are placed at the positions
recorded in the corresponding real scene, measured with the depth sensor
and the stereo views of the ZED~2i head camera and verified by
overlaying the first frame of the video on the simulated scene. As in
the real setup, the robot grasps the object and places it into the white
container (see \Cref{fig:sim_tasks}).

\textbf{Evaluation protocol.}
Each task is evaluated on 22 episodes (66 in total), one for each human
video. An episode counts as a success when the object is picked up and
placed in the container.

\section{Analysis Details}
\label{app:analysis}

\begin{table}[t]
\centering
\small
\caption{\textbf{Fr\'echet SigLIP distance to the real robot frames.}
Lower is closer. Real robot frames are from the four PnP tasks, and each
corpus is compared on all eleven tasks (as in
\Cref{fig:visual_gap}), on the four PnP tasks alone, and on the
seven tasks without robot data alone.}
\label{tab:fsd}
\setlength{\tabcolsep}{6pt}
\begin{tabular}{@{}l ccc@{}}
\toprule
Corpus & All eleven tasks & Four PnP tasks & Seven tasks w/o robot data \\
\midrule
Human            & 125.5 & 120.1 & 133.5 \\
EgoMimic         & 129.0 & 127.1 & 134.6 \\
Masquerade       & 58.2  & 50.3  & 68.8 \\
\rowcolor{oursbg} \oursb & \textbf{32.7} & \textbf{26.6} & \textbf{41.9} \\
\bottomrule
\end{tabular}
\end{table}

\begin{table}[t]
\centering
\small
\setlength{\tabcolsep}{5pt}
\caption{\textbf{Action gap, per task.} Success (\%) of replaying the action labels
of the same human videos open loop, without a policy, in a simulated
replica of the real scene with the objects at their recorded positions
(OpenArm, 22 episodes per task). Averages are in
\Cref{fig:action_gap}(c).}
\begin{tabular}{@{}l ccc c@{}}
\toprule
Labeling & PnP Bottle & PnP Box & PnP Ball & \textbf{Avg.} \\
\midrule
Retargeting \citep{qin2023anyteleop} & 63.6 & 31.8 & 9.1 & 34.8 \\
\oursb w/o wrist correction & 63.6 & \textbf{72.7} & 54.5 & 63.6 \\
\rowcolor{oursbg} \oursb & \textbf{77.3} & 63.6 & \textbf{68.2} & \textbf{69.7} \\
\bottomrule
\end{tabular}
\label{tab:action}
\end{table}

\textbf{Visual gap analysis.}
\label{app:visual_gap}
From each corpus, 5,000 frames are sampled uniformly: the human
videos, their EgoMimic and Masquerade edits, the \ours translations,
and the teleoperated robot episodes. Each frame is encoded by the
pretrained SigLIP vision tower~\citep{zhai2023siglip} of $\pi_{0.5}$ and its features are
mean-pooled over image tokens. \Cref{fig:visual_gap}(b) is a UMAP
projection fitted on the union of the five sets.
\Cref{tab:fsd} gives the Fr\'echet distance~\citep{heusel2017gans} between the SigLIP
features of each corpus and those of the real robot frames, once on the
frames of \Cref{fig:visual_gap}, once restricted to the four PnP
tasks that have robot data, and once to the seven tasks that do not.
The frames of the seven tasks without robot data lie closer to the real
frames than the edited frames of the four tasks with it, so the overlap
in \Cref{fig:visual_gap} is not a reproduction of the training
episodes.

\textbf{Action replay in simulation.}
\label{app:replay}
We replay each set of action labels in the simulation environment of
\Cref{app:sim_env}. For each human video, the labels are executed open
loop from the same initial configuration, with the objects at their
recorded positions, and each episode is evaluated as described there.
\emph{Retargeting}: the reconstructed hand \citep{zhang2025hawor} is
mapped to the robot hand \citep{qin2023anyteleop} and the arm joints are
solved by inverse kinematics from the wrist pose, the labels EgoMimic and
Masquerade train on. \emph{\ours}: the IDM labels of the synthesized
video, with and without the wrist correction of \Cref{sec:idm}. Each labeling is replayed on
the same 22 episodes of each task. On Bottle, Box, and Ball, \ours
succeeds on 17, 14, and 15 of them, \ours without the wrist correction
on 14, 16, and 12, and retargeting on 14, 7, and 2
(\Cref{tab:action}).

\textbf{Action distribution and co-training.}
\label{app:action_vocab}
\Cref{fig:action_vocab} uses right-hand actions at 20\,Hz, every frame
without subsampling. Panel~(a) shows the thumb rotation, the opposition
axis that turns the thumb toward the palm, and the mean flexion of the
other four fingers. We normalize
each joint so that the 1st and 99th percentiles of the three sources
pooled map to $-1$ and $1$, compute histogram densities over 40 bins on
$[-1.15, 1.15]$, and plot them against the corresponding joint angle in
radians. Panel~(b) is measured at the $\pi_{0.5}$ base weights before
fine-tuning, so the three methods start from the same weights and differ
only in the data they produce. Each method keeps the normalization
statistics of its own training setup. We take the gradient of the
flow-matching training loss averaged over 896 teleoperated PnP frames,
and over 880 frames of the method's human data for each out-of-domain
task. We split the episodes of each source into two halves, compute the
cosine for the two resulting pairs, and average them. The bars average
over the seven out-of-domain tasks. \emph{All params} covers the 4.1B
parameters of the model, and \emph{Action expert} covers its 0.69B
parameters together with the action and time projections.
Gradients from two halves of the same source give a cosine of 0.96 to
0.97 for teleoperation and 0.78 to 0.95 for each task, so the
differences in panel~(b) are not measurement noise. Panel~(c) uses
every frame of the right-hand actions that the policy outputs during
real robot rollouts, before filtering, in the units of panel~(a). As the
reference we sample 8,000 frames of teleoperated right-hand actions from
the four PnP tasks. Following the k-NN precision of
\citet{kynkaanniemi2019improved}, each reference frame defines a ball
whose radius is the distance to its fifth nearest reference frame, and a
rollout frame counts as within the teleoperated distribution if it falls
inside any ball. We compute this share for each policy and task and
average it over the in-domain and over the out-of-domain tasks.
Measured the same way, two halves of the teleoperated episodes give
about 73 to 80\%.

\section{Flow-Matching Objective and Sampling}
\label{app:flow}

We briefly describe conditional flow matching~\citep{lipman2023flow,liu2023rectified}, the common formulation of the policy and the three models of \ours.

Let $\mathbf{x}$ denote a clean target (\eg an action chunk or a video latent) and $\mathbf{y}$ the conditioning inputs (\eg observations and instructions) that define the conditional target distribution $p(\mathbf{x}\mid\mathbf{y})$. We construct a noisy input $\mathbf{x}^\tau$ by linearly interpolating between Gaussian noise $\boldsymbol{\epsilon}\sim\mathcal{N}(\mathbf{0},\mathbf{I})$ and $\mathbf{x}$:
\begin{equation}
    \mathbf{x}^\tau=\tau\mathbf{x}+(1-\tau)\boldsymbol{\epsilon},\qquad \tau\in[0,1],
\end{equation}
where $\tau=0$ corresponds to pure noise and $\tau=1$ to clean data. A conditional velocity network $\mathbf{v}(\mathbf{x}^\tau,\tau\mid\mathbf{y})$ is trained to predict the target velocity $\mathbf{x}-\boldsymbol{\epsilon}$:
\begin{equation}
    \mathcal{L}_{\mathrm{FM}}
    =
    \mathbb{E}_{(\mathbf{x},\mathbf{y}),\,\boldsymbol{\epsilon},\,\tau}
    \left[
        \left\|
            \mathbf{v}(\mathbf{x}^\tau,\tau\mid\mathbf{y})
            - (\mathbf{x}-\boldsymbol{\epsilon})
        \right\|_2^2
    \right].
\end{equation}
At inference, samples are generated from Gaussian noise by integrating the learned velocity from $\tau=0$ to $\tau=1$. An Euler sampler~\citep{lipman2023flow} performs this integration through repeated updates
\begin{equation}
    \mathbf{x}^{\tau+\Delta\tau}=\mathbf{x}^\tau+\Delta\tau\,\mathbf{v}(\mathbf{x}^\tau,\tau\mid\mathbf{y}),
\end{equation}
where $\Delta\tau>0$ is the step size. In \Cref{sec:method}, $f_\theta$, $g_\phi$, $h_\psi$, and $\pi_\omega$ each serve as $\mathbf{v}$, and $p_\theta$, $p_\phi$, and $p_\psi$ denote the distributions of the samples obtained in this way.

\section{Additional Results}

\begin{table}[t]
\centering
\small
\caption{\textbf{Success against collection cost, per condition.} Data
behind \Cref{fig:cost}. Each policy is trained on the four PnP
tasks alone with $N$ demonstrations per task and evaluated over 28
trials per task (112 in total). Person-hours follow
\Cref{tab:cost}: 38 teleoperated episodes or 214 human videos per
person-hour, and the equal split uses $N/2$ of each.}
\label{tab:scaling_full}
\setlength{\tabcolsep}{5pt}
\begin{tabular}{l c c c c}
\toprule
Training data & $N$ / task & Person-hours & Successes / 112 & SR (\%) \\
\midrule
Teleop only            & 10  & 1.05 & 21 & 18.8 \\
                       & 30  & 3.16 & 39 & 34.8 \\
                       & 50  & 5.26 & 53 & 47.3 \\
\midrule
Human video only       & 10  & 0.19 & 14 & 12.5 \\
                       & 30  & 0.56 & 28 & 25.0 \\
                       & 50  & 0.93 & 40 & 35.7 \\
                       & 100 & 1.87 & 58 & 51.8 \\
\midrule
Teleop + human (1:1)   & 10  & 0.62 & 19 & 17.0 \\
                       & 30  & 1.86 & 40 & 35.7 \\
                       & 50  & 3.10 & 56 & 50.0 \\
                       & 100 & 6.20 & 80 & \textbf{71.4} \\
\bottomrule
\end{tabular}
\end{table}

\begin{table}[t]
\centering
\caption{\textbf{Data collection cost.} Teleoperation occupies two people,
so one person-hour yields 14\,min of robot demonstration against
45\,min of egocentric human video. Because a person also performs
each task faster than a teleoperated robot, the gap in episode count is
5--6$\times$.}
\label{tab:cost}
\begin{tabular}{l cc}
\toprule
 & Robot Teleoperation & Human Ego Video \\
\midrule
People per session               & 2 & 1 \\
Demonstration footage / person-hour & 14\,min & 45\,min \\
Episodes / person-hour           & 38 & 214 \\
Hardware                         & Robot + teleoperation system & Fixed camera \\
Trained operator required        & Yes & No \\
Occupies robot                   & Yes & No \\
Bimanual overhead                & High (two arms to operate) & None \\
\bottomrule
\end{tabular}
\end{table}

\subsection{Data Collection Cost}
\label{app:cost}
\Cref{tab:cost} lists the cost model behind the person-hour axis of
\Cref{fig:cost}, and \Cref{tab:scaling_full} the eleven
policies plotted there. Teleoperation occupies two people and yields
about 14 minutes of demonstration per person-hour, whereas a single person
records about 45 minutes of human video in the same time and completes
each task faster, so the gap in episode count (38 against 214 per
person-hour) is larger than the gap in footage. The teleop-only policy
with 50 demonstrations per task is the \emph{Teleop only} policy of
\Cref{tab:main_openarm}. The other ten policies are trained for this
analysis on the four PnP tasks alone, so their values are not those of
\Cref{tab:main_openarm}, whose policies also train on the seven tasks
without robot demonstrations.

\subsection{Per-Task Results}
\label{app:extra}

\begin{table}[t]
\centering
\caption{
\textbf{Per-task results behind \Cref{tab:main_real}.}
Success rates (\%) over 28 trials per task using $\pi_{0.5}$ on
OpenArm and RB-Y1.
Robot demonstrations are available only for the PnP tasks in (a).
``+'' rows add the human videos converted by each method to the
teleoperated demonstrations, and the \oursb{} only rows use the synthesized
episodes alone. EgoMimic and Masquerade are deployed with their
test-time mask and overlay, whereas \ours uses the raw camera view.
}
\label{tab:main_full}

\begin{subtable}{\textwidth}
\centering
\caption{Tasks with robot demonstrations}
\label{tab:main_full_a}
\footnotesize
\setlength{\tabcolsep}{4pt}
\begin{tabular}{l cc cccc c}
\toprule
& \multicolumn{2}{c}{Data}
& \multicolumn{4}{c}{In-domain tasks} & \\
\cmidrule(lr){2-3}
\cmidrule(lr){4-7}
Method & Teleop & Human
& PnP Bottle & PnP Doll & PnP Box & PnP Ball & Avg. \\
\midrule
\multicolumn{8}{l}{\textit{OpenArm}} \\
Teleop only
& \textcolor{SJViolet}{\cmark} & \textcolor{SJRed}{\xmark}
& 71.4 & 39.3 & 42.9 & 35.7 & 47.3 \\
+ EgoMimic~\citep{kareer2025egomimic}
& \textcolor{SJViolet}{\cmark} & \textcolor{SJViolet}{\cmark}
& 64.3 & 50.0 & 42.9 & 42.9 & 50.0 \\
+ Masquerade~\citep{lepert2025masquerade}
& \textcolor{SJViolet}{\cmark} & \textcolor{SJViolet}{\cmark}
& \textbf{92.9} & 42.9 & \textbf{71.4} & 64.3 & 67.9 \\
+ \oursb{}
& \textcolor{SJViolet}{\cmark} & \textcolor{SJViolet}{\cmark}
& \textbf{92.9} & \textbf{71.4} & \textbf{71.4}
& \textbf{67.9} & \textbf{75.9} \\
\midrule
\oursb{} only
& \textcolor{SJRed}{\xmark} & \textcolor{SJViolet}{\cmark}
& 82.1 & 57.1 & 50.0 & 39.3 & 57.1 \\
\midrule
\multicolumn{8}{l}{\textit{RB-Y1}} \\
Teleop only
& \textcolor{SJViolet}{\cmark} & \textcolor{SJRed}{\xmark}
& 82.1 & 64.3 & 75.0 & 57.1 & 69.6 \\
+ EgoMimic
& \textcolor{SJViolet}{\cmark} & \textcolor{SJViolet}{\cmark}
& 67.9 & 71.4 & 64.3 & 50.0 & 63.4 \\
+ Masquerade
& \textcolor{SJViolet}{\cmark} & \textcolor{SJViolet}{\cmark}
& 67.9 & 57.1 & 67.9 & 60.7 & 63.4 \\
+ \oursb{}
& \textcolor{SJViolet}{\cmark} & \textcolor{SJViolet}{\cmark}
& \textbf{89.3} & \textbf{75.0} & \textbf{85.7} & \textbf{67.9} & \textbf{79.5} \\
\midrule
\oursb{} only
& \textcolor{SJRed}{\xmark} & \textcolor{SJViolet}{\cmark}
& 71.4 & 64.3 & 75.0 & 32.1 & 60.7 \\
\bottomrule
\end{tabular}
\end{subtable}

\medskip

\begin{subtable}{\textwidth}
\centering
\caption{Tasks without task-specific robot demonstrations}
\label{tab:main_full_b}
\scriptsize
\setlength{\tabcolsep}{4pt}
\begin{tabular}{l cc ccccccc c}
\toprule
& \multicolumn{2}{c}{Data}
& Novel Object & Novel Goal
& \multicolumn{2}{c}{Novel Skill}
& \multicolumn{3}{c}{Bimanual} & \\
\cmidrule(lr){2-3}
\cmidrule(lr){4-4}
\cmidrule(lr){5-5}
\cmidrule(lr){6-7}
\cmidrule(lr){8-10}
Method & Teleop & Human
& \shortstack{PnP\\Coffee Pot}
& \shortstack{PnP Bottle\\to Shelf}
& Pouring
& Brushing
& \shortstack{Box\\Convey}
& \shortstack{Air\\Fryer}
& \shortstack{PnP Ball\\in Cup}
& Avg. \\
\midrule
\multicolumn{11}{l}{\textit{OpenArm}} \\
Teleop only
& \textcolor{SJViolet}{\cmark} & \textcolor{SJRed}{\xmark}
& 14.3 & 7.1 & 0.0 & 0.0 & 0.0 & 0.0 & 0.0 & 3.1 \\
+ EgoMimic
& \textcolor{SJViolet}{\cmark} & \textcolor{SJViolet}{\cmark}
& 14.3 & 0.0 & 7.1 & 14.3 & 0.0 & 0.0 & 0.0 & 5.1 \\
+ Masquerade
& \textcolor{SJViolet}{\cmark} & \textcolor{SJViolet}{\cmark}
& 21.4 & 7.1 & 7.1 & 21.4 & 0.0 & 0.0 & 0.0 & 8.2 \\
+ \oursb{}
& \textcolor{SJViolet}{\cmark} & \textcolor{SJViolet}{\cmark}
& \textbf{64.3} & \textbf{89.3} & \textbf{85.7}
& \textbf{82.1} & \textbf{78.6} & \textbf{60.7}
& \textbf{50.0} & \textbf{73.0} \\
\oursb{} only
& \textcolor{SJRed}{\xmark} & \textcolor{SJViolet}{\cmark}
& 50.0 & 57.1 & 64.3 & 53.6 & 57.1 & 35.7 & 21.4 & 48.5 \\
\midrule
\multicolumn{11}{l}{\textit{RB-Y1}} \\
Teleop only
& \textcolor{SJViolet}{\cmark} & \textcolor{SJRed}{\xmark}
& 7.1 & 0.0 & 0.0 & 0.0 & 0.0 & 0.0 & 0.0 & 1.0 \\
+ EgoMimic
& \textcolor{SJViolet}{\cmark} & \textcolor{SJViolet}{\cmark}
& 14.3 & 0.0 & 10.7 & 10.7 & 0.0 & 0.0 & 0.0 & 5.1 \\
+ Masquerade
& \textcolor{SJViolet}{\cmark} & \textcolor{SJViolet}{\cmark}
& 21.4 & 7.1 & 10.7 & 14.3 & 0.0 & 0.0 & 0.0 & 7.7 \\
+ \oursb{}
& \textcolor{SJViolet}{\cmark} & \textcolor{SJViolet}{\cmark}
& 60.7 & \textbf{67.9} & \textbf{75.0} & 39.3 & \textbf{32.1} & 0.0 & \textbf{53.6} & \textbf{46.9} \\
\oursb{} only
& \textcolor{SJRed}{\xmark} & \textcolor{SJViolet}{\cmark}
& \textbf{64.3} & 60.7 & 60.7 & \textbf{46.4} & 28.6 & 0.0 & 21.4 & 40.3 \\
\bottomrule
\end{tabular}
\end{subtable}
\end{table}

\Cref{tab:main_full} gives the per-task results for the main experiments.
\Cref{tab:backbone_full} gives the per-task results for the two policy
backbones, $\pi_{0.5}$ and RLDX-1-Img~\citep{rldx2026}, behind
\Cref{tab:backbone}. \Cref{tab:ablation_full} gives the per-task
results behind \Cref{tab:ablation}.

\subsection{Ablated Variants}
\label{app:ablations}
\begin{figure}[t]
\centering
\includegraphics[width=\textwidth]{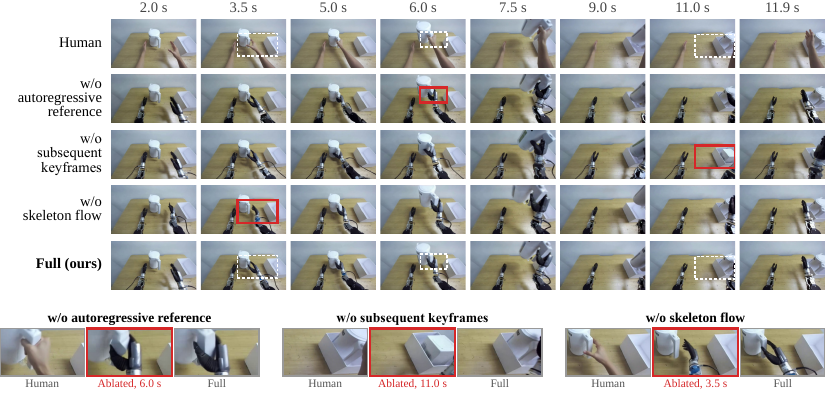}
\caption{\textbf{Failure cases of the ablated variants.}
Human video and videos synthesized by the variants in
\Cref{tab:ablation} at the same time steps. Boxed regions are
enlarged at representative failure points.}
\label{fig:ablation_fail}
\end{figure}

\Cref{fig:ablation_fail} shows failure cases of the ablated
variants on \emph{PnP Coffee Pot}. Without the autoregressive reference,
each keyframe is translated independently, resulting in inconsistent
grasp configurations: the fingers wrap the handle from below at
5.0\,s and reach behind it at 6.0\,s, causing visible hand jitter
between keyframes. In contrast, the full model maintains a consistent
grasp throughout.

Without subsequent keyframes, the video is generated from the first keyframe
alone. Although the hand still follows the skeleton, the coffee pot
changes into a box inside the container at 11.0\,s. Without skeleton
flow, the motion between keyframes deviates from the demonstrator's
trajectory, causing the wrist and hand to lose the configuration
required to grasp the handle at 3.5\,s.

\subsection{Qualitative Results}
\label{app:qual}

\begin{figure}[t]
\centering
\setlength{\embw}{0.2\textwidth}
\setlength{\tabcolsep}{1pt}
\renewcommand{\arraystretch}{0.8}
\newcommand{\et}[2]{\thumbat{\embw}{embodiments-adj/#1}{#2}}
{\small
\begin{tabular}{@{}cccc@{}}
Human & OpenArm & RB-Y1 & GR-1 \\[1pt]
\et{ep1-human.jpg}{h1} & \et{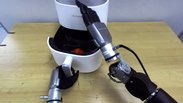}{o1} & \et{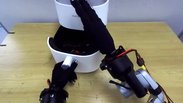}{r1} & \et{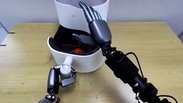}{g1} \\
\et{ep2-human.jpg}{h2} & \et{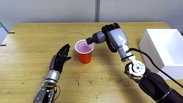}{o2} & \et{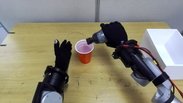}{r2} & \et{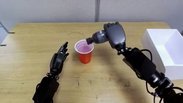}{g2} \\
\et{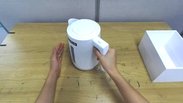}{h3} & \et{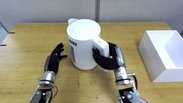}{o3} & \et{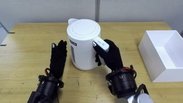}{r3} & \et{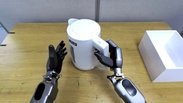}{g3} \\
\end{tabular}}
\caption{\textbf{One human video, three robots.} Human videos (left)
translated into the OpenArm, the RB-Y1, and the GR-1 by a single set of
models, and only the embodiment changes.}
\label{fig:embodiments}
\end{figure}

\Cref{fig:embodiments} shows human videos translated into the OpenArm,
the RB-Y1, and the GR-1 by a single set of models.

\begin{table}[t]
\centering
\scriptsize
\caption{\textbf{Policy backbone, per task.} Success rate (\%) on the
OpenArm for each backbone, trained on the teleoperated episodes alone
and with the episodes synthesized by \ours.}
\label{tab:backbone_full}
\setlength{\tabcolsep}{2pt}
\begin{adjustbox}{max width=\textwidth}
\begin{tabular}{@{}l l cccc c ccccccc c@{}}
\toprule
& & \multicolumn{5}{c}{In-domain} & \multicolumn{8}{c}{Out-of-domain} \\
\cmidrule(lr){3-7}\cmidrule(lr){8-15}
Backbone & Data & \shortstack{PnP\\Bottle} & \shortstack{PnP\\Doll} & \shortstack{PnP\\Box} & \shortstack{PnP\\Ball} & Avg.
& \shortstack{PnP\\Coffee Pot} & \shortstack{PnP Bottle\\to Shelf} & Pouring & Brushing & \shortstack{Box\\Convey} & \shortstack{Air\\Fryer} & \shortstack{PnP Ball\\in Cup} & Avg. \\
\midrule
$\pi_{0.5}$ & Teleop only & 71.4 & 39.3 & 42.9 & 35.7 & 47.3 & 14.3 & 7.1 & 0.0 & 0.0 & 0.0 & 0.0 & 0.0 & 3.1 \\
            & + \oursb    & \textbf{92.9} & \textbf{71.4} & \textbf{71.4} & \textbf{67.9} & \textbf{75.9} & \textbf{64.3} & \textbf{89.3} & \textbf{85.7} & \textbf{82.1} & \textbf{78.6} & \textbf{60.7} & \textbf{50.0} & \textbf{73.0} \\
\midrule
RLDX-1-Img  & Teleop only & 50.0 & \textbf{100.0} & 57.1 & 7.1 & 53.6 & 0.0 & 0.0 & 0.0 & 0.0 & 0.0 & 0.0 & 0.0 & 0.0 \\
            & + \oursb    & \textbf{57.1} & 57.1 & \textbf{64.3} & \textbf{64.3} & \textbf{60.7} & \textbf{28.6} & \textbf{21.4} & \textbf{42.9} & \textbf{32.1} & \textbf{64.3} & \textbf{21.4} & \textbf{71.4} & \textbf{40.3} \\
\bottomrule
\end{tabular}
\end{adjustbox}
\end{table}

\begin{table}[t]
\centering
\scriptsize
\caption{\textbf{Component ablation, per task.} Success rate (\%) on the
OpenArm with $\pi_{0.5}$ for the variants of \Cref{tab:ablation}.}
\label{tab:ablation_full}
\setlength{\tabcolsep}{2pt}
\begin{adjustbox}{max width=\textwidth}
\begin{tabular}{@{}l cccc c ccccccc c@{}}
\toprule
& \multicolumn{5}{c}{In-domain} & \multicolumn{8}{c}{Out-of-domain} \\
\cmidrule(lr){2-6}\cmidrule(lr){7-14}
Variant & \shortstack{PnP\\Bottle} & \shortstack{PnP\\Doll} & \shortstack{PnP\\Box} & \shortstack{PnP\\Ball} & Avg.
& \shortstack{PnP\\Coffee Pot} & \shortstack{PnP Bottle\\to Shelf} & Pouring & Brushing & \shortstack{Box\\Convey} & \shortstack{Air\\Fryer} & \shortstack{PnP Ball\\in Cup} & Avg. \\
\midrule
\rowcolor{oursbg} \oursb (full) & 92.9 & 71.4 & 71.4 & 67.9 & 75.9 & 64.3 & 89.3 & 85.7 & 82.1 & 78.6 & 60.7 & 50.0  & 73.0 \\
\midrule
w/o Autoregressive reference & 85.7 & 64.3 & 64.3 & 60.7 & 68.8 & 57.1 & 82.1 & 78.6 & 75.0 & 71.4 & 53.6 & 46.4  & 66.3 \\
w/o Subsequent keyframes & 53.6 & 39.3 & 39.3 & 39.3 & 42.9 & 21.4 & 17.9 & 53.6 & 60.7 & 53.6 & 39.3 & 28.6  & 39.3 \\
w/o Hand skeleton flow & 50.0 & 39.3 & 39.3 & 39.3 & 42.0 & 10.7 & 42.9 & 42.9 & 28.6 & 14.3 & 7.1 & 3.6  & 21.4 \\
w/o IDM wrist correction & 50.0 & 64.3 & 78.6 & 67.9 & 65.2 & 35.7 & 53.6 & 46.4 & 75.0 & 71.4 & 32.1 & 21.4 & 48.0 \\
\bottomrule
\end{tabular}
\end{adjustbox}
\end{table}

\begin{figure}[t]
\centering
\begin{subfigure}{\textwidth}
  \centering
  \includegraphics[width=\textwidth]{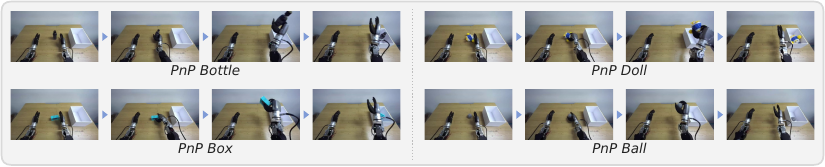}
  \caption{In-domain tasks}
\end{subfigure}\\[4pt]
\begin{subfigure}{\textwidth}
  \centering
  \includegraphics[width=\textwidth]{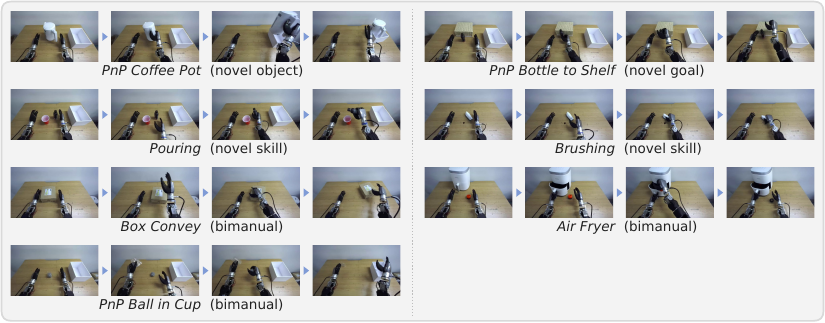}
  \caption{Out-of-domain tasks}
\end{subfigure}
\caption{\textbf{Rollouts on all eleven tasks on the OpenArm}, with the
policy trained on the teleoperated and synthesized episodes: (a) the
four tasks with robot demonstrations and (b) the seven with human video
only.}
\label{fig:rollouts_all}
\end{figure}

\begin{figure}[t]
\centering
\begin{subfigure}{\textwidth}
  \centering
  \includegraphics[width=\textwidth]{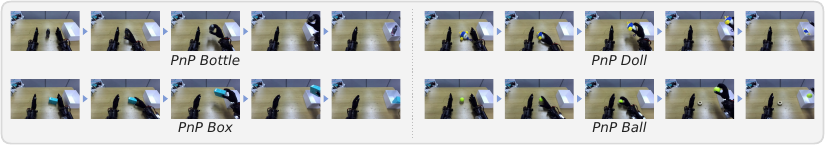}
  \caption{In-domain tasks}
\end{subfigure}\\[4pt]
\begin{subfigure}{\textwidth}
  \centering
  \includegraphics[width=\textwidth]{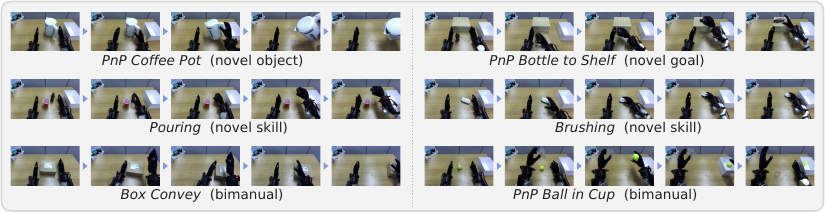}
  \caption{Out-of-domain tasks}
\end{subfigure}
\caption{\textbf{Rollouts on the RB-Y1}, with the policy trained on the
teleoperated and synthesized episodes: (a) the four tasks with robot
demonstrations and (b) the tasks with human video only. \emph{Air Fryer}
is omitted because no policy completes it on the RB-Y1 (0\% success in
\Cref{tab:main_full}).}
\label{fig:rollouts_rby1}
\end{figure}

\begin{figure}[t]
\centering
\figorbox{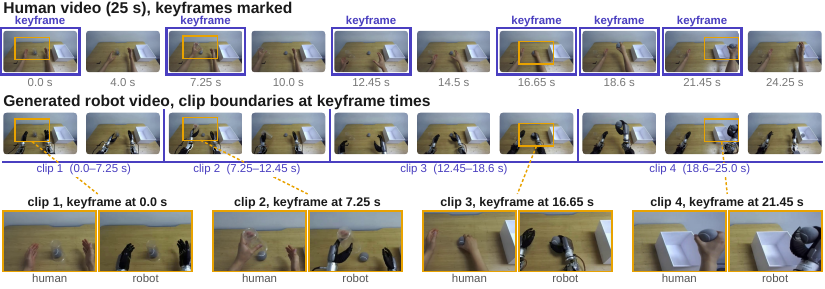}{\textwidth}{4.5cm}
\caption{\textbf{Long-horizon synthesis.} A 25-second human video (top)
with its selected keyframes, and the generated robot video (bottom) with
the clip boundaries at the keyframe times. Adjacent clips share the keyframe
at their boundary, so robot and objects stay consistent across
boundaries.}
\label{fig:longvideo}
\end{figure}

\subsubsection{Rollouts on OpenArm}
In \Cref{fig:rollouts_all}, we show all rollouts on the OpenArm from the
policy trained on both teleoperated and synthesized episodes.
\Cref{fig:rollouts_all}(a) shows the four tasks with robot demonstrations,
while \Cref{fig:rollouts_all}(b) shows the seven tasks with human video
only.

\subsubsection{Rollouts on RB-Y1}
In \Cref{fig:rollouts_rby1}, we show all rollouts on the RB-Y1 from the
policy trained on both teleoperated and synthesized episodes.
\Cref{fig:rollouts_rby1}(a) shows the four tasks with robot demonstrations,
while \Cref{fig:rollouts_rby1}(b) shows the six tasks with human video
only, without \emph{Air Fryer}.

No method completes \emph{Air Fryer} on the RB-Y1 (\Cref{tab:main_full}).
The task requires inserting the fingers between the handle and the
fryer body before closing, a fine-grained grasping motion not observed
in the teleoperated data and beyond the resolution of the IDM labels.
The longer, multi-joint fingers of the WujiHand2 also tend to push the
fryer away when closing in front of the handle, unlike the shorter
1-DoF fingers of the Inspire RH56F1.

\subsubsection{Long-Horizon Synthesis}
\Cref{fig:longvideo} shows a 25-second human video and the
corresponding long-horizon robot video generated from its selected
keyframes. The generated video is divided into clips at the keyframe
times, and adjacent clips share the keyframe at their boundary. This preserves the consistency of the robot and
objects across clip boundaries.

\begin{figure}[t]
\centering
\includegraphics[width=\textwidth]{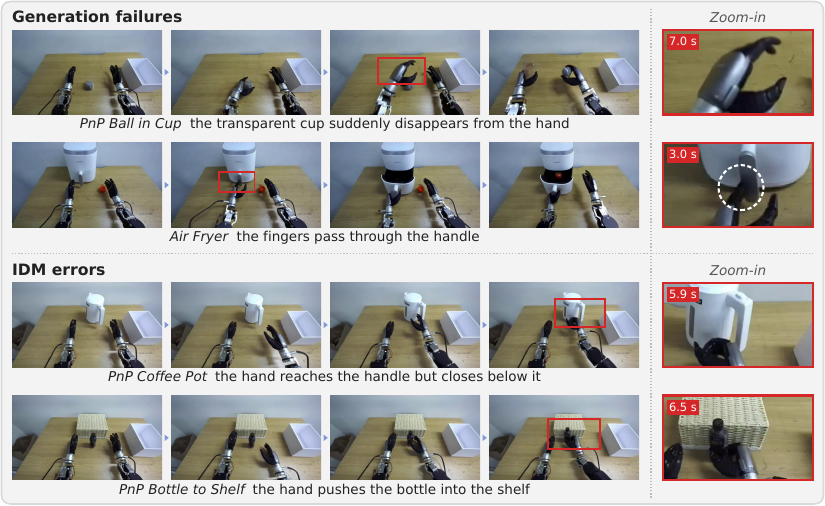}
\caption{\textbf{Failure cases.} The red box in each row marks the
region enlarged on the right. \emph{Generation failures} (top):
synthesized videos in which the generative models do not preserve the
objects. On \emph{PnP Ball in Cup} the transparent cup suddenly
disappears from the hand as it is lifted, leaving the ball on the
table. On \emph{Air Fryer} the fingers pass through the handle instead
of closing around it (dashed circle). \emph{IDM errors} (bottom): policy
rollouts that fail because the action labels of the synthesized
episodes place the hand imprecisely. On \emph{PnP Coffee Pot} the hand
reaches the handle but closes below it, and on \emph{PnP Bottle to
Shelf} it closes low on the bottle and pushes it against the shelf
instead of lifting it onto it.}
\label{fig:failures}
\end{figure}

\end{document}